\documentclass[a4paper,fleqn]{cas-dc}

\usepackage[numbers,sort&compress]{natbib}

\def\tsc#1{\csdef{#1}{\textsc{\lowercase{#1}}\xspace}}
\tsc{WGM}
\tsc{QE}
\usepackage{booktabs}
\usepackage{multirow}
\usepackage{graphicx}
\usepackage{placeins}
\usepackage{algpseudocode}
\usepackage[ruled,vlined]{algorithm2e}
\usepackage[table]{xcolor}
\usepackage{xcolor}
\usepackage{bm}
\usepackage{silence}
\usepackage{threeparttable}
\usepackage{subcaption}
\usepackage{siunitx}
\usepackage{afterpage}
\usepackage{xcolor}

\usepackage{colortbl}
\definecolor{bestblue}{RGB}{210, 230, 255}

\usepackage{longtable}   % For multi-page tables
\usepackage{makecell}    % For \makecell
\usepackage{ltxtable}

\usepackage{soul}

\ifdefined\pdfsuppresswarningpagegroup
\fi

\makeatletter
\let\font@warning\@gobble
\makeatother
\hypersetup{hypertexnames=false}

\ExplSyntaxOn
\RenewDocumentEnvironment { Abstract } { o }
{
  \group_begin:
  \IfNoValueTF { #1 } { }
    { \tex_gdef:D \abstractname { #1 } }
  \parindent 0pt
  \box_if_empty:NTF \g_stm_key_box
    { \leftskip = .35 \textwidth }
    {
      \dim_gset:Nn \l_tmpa_dim { \box_ht:N \g_stm_key_box }
      \dim_gadd:Nn \l_tmpa_dim { \box_dp:N \g_stm_key_box }
      \leftskip .35\textwidth
      \hspace*{-.35 \textwidth }
      \noindent\hbox_to_wd:nn { 0pt } { \box \g_stm_key_box \hss }
      \skip_vertical:n { - \l_tmpa_dim }
    }
  \noindent \abstractname \par
  \skip_vertical:n { -4pt }
  \noindent \rule{.65\textwidth}{.2pt}\par \footnotesize
  \ignorespaces \everypar { \parindent=1.5em }
}
{ \par \group_end: }
\ExplSyntaxOff

\definecolor{JournalBlue}{HTML}{1F4E79}
\definecolor{JournalLightBlue}{HTML}{EAF1F7}
\definecolor{bestblue}{HTML}{E7F0F7}
\definecolor{secondred}{HTML}{F7E3E5}
\definecolor{improvegreen}{HTML}{E8F2EC}
\definecolor{regressred}{HTML}{F7E3E5}
\definecolor{neutralgray}{HTML}{F3F3F3}
\definecolor{deltaup}{HTML}{478B68}
\definecolor{deltadown}{HTML}{B56B75}
\definecolor{rankednavy}{HTML}{17365D}
\definecolor{biasgood}{HTML}{E8F2EC}
\definecolor{biasbad}{HTML}{F7E3E5}
\makeatletter
\def\InlineResultBreak{\@ifnextchar[{\InlineResultBreak@i}{\nobreak\hspace{0.18em}}}
\def\InlineResultBreak@i[#1]{\nobreak\hspace{0.18em}}
\newcommand{\InlineResult}[1]{%
  \begingroup
  \let\\\InlineResultBreak
  #1%
  \endgroup}

\newcommand{\CompactDeltaResult}[2]{%
  \begingroup
  \setbox\z@=\hbox{#1}%
  \setbox\tw@=\hbox{{\fontsize{5.0}{5.0}\selectfont #2}}%
  \dimen@=\wd\z@
  \ifdim\wd\tw@>\dimen@ \dimen@=\wd\tw@\fi
  \vtop{%
    \offinterlineskip
    \hbox to \dimen@{\hfil\box\z@\hfil}%
    \vskip0.3pt
    \hbox to \dimen@{\hfil\box\tw@\hfil}}%
  \endgroup}
\makeatother
\algrenewcommand\algorithmicrequire{\textbf{Input:}}
\algrenewcommand\algorithmicensure{\textbf{Output:}}

\begin{document}
\let\WriteBookmarks\relax
% Float settings for a figure- and table-rich double-column manuscript.
\setcounter{topnumber}{4}
\setcounter{bottomnumber}{2}
\setcounter{totalnumber}{6}
\setcounter{dbltopnumber}{4}
\renewcommand{\topfraction}{0.95}
\renewcommand{\bottomfraction}{0.85}
\renewcommand{\textfraction}{0.05}
\renewcommand{\floatpagefraction}{0.70}
\renewcommand{\dbltopfraction}{0.95}
\renewcommand{\dblfloatpagefraction}{0.60}
\hbadness=10000
\hfuzz=0.5pt
\emergencystretch=2em

\shorttitle{\hspace{0pt}}    

\shortauthors{\hspace{0pt}}  

\title [mode = title]{OncoVision: Integrating Mammography and Clinical Data through Attention-Driven Multimodal AI for Enhanced Breast Cancer Diagnosis} 

% % Title footnote mark
% % eg: \tnotemark[1]

% % Title footnote 1.
% % eg: \tnotetext[1]{Title footnote text}

% =======================================
% AUTHORS
% =======================================

% ============================================================================
% AUTHORS
% ============================================================================

\author[1,3]{Istiak Ahmed}
\ead{istiak.ahmed1@northsouth.edu}
\credit{Conceptualization, Data Curation, Formal Analysis, Investigation, Methodology, Software, Validation, Visualization, Writing – original draft, Writing – review and editing}

\author[2,3]{Galib Ahmed}
\ead{galib.ahmed.251@northsouth.edu}
\credit{Conceptualization, Data Curation, Formal Analysis, Investigation, Methodology, Software, Validation, Visualization, Writing – original draft, Writing – review and editing}

\author[1,3]{Kazi Shahriar Sanjid}
\ead{kazi.sanjid@northsouth.edu}
\credit{Data Curation, Formal Analysis, Methodology, Software, Validation, Visualization, Writing – review and editing}

\author[4,3]{Md. Tanzim Hossain}
\ead{tanzim.hossain@fau.de}
\credit{Formal Analysis, Software, Validation, Visualization, Writing – review and editing}

\author[1,3]{Md. Nishan Khan}
\ead{nishan.khan@northsouth.edu}
\credit{Data Curation, Methodology, Software, Validation}

\author[1,3]{Md. Misbah Khan}
\ead{misbah.khan@northsouth.edu}
\credit{Data Curation, Methodology, Software, Visualization}

\author[5]{Md. Arifur Rahman}
\ead{drarifur@bshl.com.bd}
\credit{Formal Analysis, Resources, Validation, Visualization}

\author[6]{Sheikh Anisul Haque}
\ead{dr.anisul@bshl.com.bd}
\credit{Formal Analysis, Resources, Validation}

\author[7]{Sharmin Akhtar Rupa}
\ead{dr.sharminakhtarrupa@yahoo.com}
\credit{Formal Analysis, Resources, Validation, Visualization}

\author[8]{Mohammed Mejbahuddin Mia}
\ead{drmejbah2013@gmail.com}
\credit{Formal Analysis, Resources, Validation}

\author[9]{Mahmud Hasan Mostofa Kamal}
\ead{mkamal@bsmmu.edu.bd}
\credit{Formal Analysis, Resources, Project Administration, Validation, Writing – review and editing}

\author[10]{Md. Mostafa Kamal Sarker}
\ead{md.sarker@ndm.ox.ac.uk}
\credit{Conceptualization, Formal Analysis, Methodology, Validation, Writing – review and editing}

\author[2,3]{M. Monir Uddin}
\cormark[1]
\ead{monir.uddin@northsouth.edu}
\credit{Conceptualization, Formal Analysis, Funding acquisition, Methodology, Project administration, Resources, Supervision, Writing – review and editing}

\cortext[1]{Corresponding author}

% ============================================================================
% AFFILIATIONS
% ============================================================================

\affiliation[1]{%
    addressline={Department of Electrical and Computer Engineering},
    organization={North South University},
    city={Dhaka},
    postcode={1229},
    country={Bangladesh}
}

\affiliation[2]{%
    addressline={Department of Mathematics and Physics},
    organization={North South University},
    city={Dhaka},
    postcode={1229},
    country={Bangladesh}
}

\affiliation[3]{%
    addressline={Big-Matrix Lab, Department of Mathematics and Physics},
    organization={North South University},
    city={Dhaka},
    postcode={1229},
    country={Bangladesh}
}

\affiliation[4]{%
    addressline={Department of Data Science},
    organization={Friedrich-Alexander University},
    city={Erlangen},
    postcode={91054},
    country={Germany}
}

\affiliation[5]{%
    addressline={Department of Oncology \& Radiotherapy},
    organization={Bangladesh Specialized Hospital},
    city={Dhaka},
    postcode={1207},
    country={Bangladesh}
}

\affiliation[6]{%
    addressline={Department of Transfusion Medicine},
    organization={Bangladesh Specialized Hospital},
    city={Dhaka},
    postcode={1207},
    country={Bangladesh}
}

\affiliation[7]{%
    addressline={Department of Radiology \& Imaging},
    organization={Popular Medical College},
    city={Dhaka},
    postcode={1205},
    country={Bangladesh}
}

\affiliation[8]{%
    addressline={Department of Transfusion Medicine},
    organization={Khwaja Yunus Ali Medical College \& Hospital},
    city={Sirajganj},
    postcode={6751},
    country={Bangladesh}
}

\affiliation[9]{%
    addressline={Department of Radiology and Imaging},
    organization={Bangladesh Medical University},
    city={Dhaka},
    postcode={1000},
    country={Bangladesh}
}

\affiliation[10]{%
    addressline={Nuffield Department of Medicine, Old Road Campus},
    organization={University of Oxford},
       city={Oxford},
    postcode={OX3 7LF},
    country={United Kingdom}
}


%\fntext[1]{Draft author information.}



\begin{abstract}
OncoVision is a privileged-information training framework that uses mammography images, clinical features, and radiomic descriptors during training and performs inference from mammographic images alone, for AI-assisted mammographic interpretation and structured diagnostic assessment. Employing an attention-based encoder--decoder backbone, it jointly segments four regions of interest masses, calcifications, axillary findings, and breast tissue with accuracy exceeding the nnU-Net baseline and predicts ten structured clinical features, including mass morphology, calcification type, ACR breast density, and BI-RADS category. To integrate learned imaging representations, explicit radiomic descriptors, and clinical information during training, we developed two late-fusion strategies Independent and Dependent that use complementary multimodal data to improve diagnostic precision and have the potential to reduce inter-observer variability. Radiomic features extracted from predicted masks provide explicit shape, intensity, and texture descriptors that complement the learned CNN representations. We evaluated OncoVision through a retrospective multi-reader, multi-case study involving six board-certified radiologists across three experience levels, assessing diagnostic confidence, reading time, and segmentation accuracy with and without AI assistance. In a paired reader-assistance evaluation, OncoVision was associated with higher diagnostic confidence for junior and senior radiologists, reduced reading time by up to 61\%, and achieved segmentation accuracy comparable to or exceeding that of radiologists for mass lesions. To assess real-world utility, we operationalized OncoVision as a secure, user-friendly web application, now deployed at a partner hospital, which generates structured reports with dual-confidence scoring and attention-weighted visualizations for real-time diagnostic support, clinician trust, and medical education. The platform is designed for integration into clinical workflows, with the goal of supporting screening access in underprivileged regions such as rural South Asia. By combining accurate segmentation with clinical intuition, OncoVision advances AI-assisted mammographic interpretation, offering a scalable and accessible approach to support earlier and more consistent image interpretation.
\end{abstract}

% Keywords

\begin{NoHyper}
\maketitle
\end{NoHyper}

% Main text

% Use the style of numbering in square brackets.
% If nothing is used, default style will be taken.
%\begin{enumerate}[a)]
%\item 
%\item 
%\item 
%\end{enumerate}  

%\begin{itemize}
%\item 
%\item 
%\item 
%\end{itemize}  

%\begin{description}
%\item[]
%\item[] 
%\item[] 
%\end{description}  

\section{Introduction}\label{sec:intro}

Breast cancer remains the leading cause of cancer-related mortality among women worldwide, with over 2.3 million new cases and 685,000 deaths reported annually~\citep{ferlay2020global, who2020cancer}. Early detection through mammography is the cornerstone of effective intervention, substantially improving survival rates and reducing treatment costs~\citep{siegel2023cancer, 10.1259/bjr.20211033}. However, mammographic interpretation remains challenging: low-contrast microcalcifications, isodense masses with spiculated margins, and complex Breast Imaging Reporting and Data System (BI-RADS) categories 4 and 5 frequently evade detection or are misinterpreted~\citep{elmore2009variability, lehman2017diagnostic, birdwell2001mammographic}. Inter-observer variability among radiologists remains substantial, and in resource-limited settings, where expert radiologists and advanced imaging equipment are scarce, these diagnostic disparities are further exacerbated~\citep{pace2014challenges, perry2008breast, chen2019healthcare}.

Artificial intelligence (AI) has emerged as a promising tool to augment mammographic interpretation by automating detection, segmentation, and risk stratification~\citep{hosny2018artificial, topol2019high, shen2021ai}. Yet most existing AI systems address isolated tasks: detecting a single abnormality, segmenting one region of interest, or predicting one clinical feature~\citep{kelly2019key, kaushal2020geographic}. Such fragmented approaches do not reflect the structured, multi-faceted reasoning that radiologists employ in clinical practice, where a single mammogram yields a comprehensive report encompassing mass morphology, calcification distribution, axillary findings, breast density, and a final BI-RADS assessment~\citep{dorsi2013acr, sickles2013birads}. Clinically useful AI must therefore move beyond isolated task optimization toward interpretable pipelines that mirror the radiologist's diagnostic workflow and can be integrated into hospital systems~\citep{reyes2020interpretability, tonekaboni2019clinicians}.

We hypothesized that an integrated multimodal AI pipeline, one that jointly segments multiple anatomical regions and predicts a structured set of clinical features, could improve diagnostic accuracy, consistency, and efficiency, particularly for non-expert radiologists and in resource-constrained settings. To test this hypothesis, we developed OncoVision, a privileged-information training framework that uses clinical data during training and performs inference from mammographic images and internally derived radiomic descriptors alone, through concatenation-based late fusion~\citep{baur2025multimodal}. OncoVision simultaneously segments four regions of interest (masses, calcifications, axillary findings, and breast tissue) and predicts ten structured clinical features, including mass morphology, calcification distribution, ACR breast density, and BI-RADS category. Radiomic features extracted from the predicted masks provide explicit shape, intensity, and texture descriptors that improve clinical feature prediction during training. We evaluated OncoVision through a retrospective multi-reader, multi-case study involving six board-certified radiologists across three experience levels, assessing diagnostic confidence, reading time, and segmentation accuracy with and without AI assistance~\citep{kim2025effect}.

A key design question in multimodal AI is how best to integrate imaging and tabular clinical data during training. We therefore investigated two late-fusion strategies. The Independent variant concatenates raw clinical features directly with imaging-derived bottleneck representations and radiomic descriptors, preserving the original clinical input without learned transformation. The Dependent variant employs a dedicated tabular encoder to learn a compact representation of clinical features before fusion, allowing richer interactions between modalities. This design enables us to evaluate two training regimes: one relying on minimally processed clinical inputs, and one using learned tabular representations. Both variants proved viable, with the Dependent variant achieving higher accuracy on most tasks.

Our hypothesis was justified through a multi-faceted evaluation strategy. First, we established technical validity by benchmarking OncoVision against the nnU-Net~\citep{isensee2021nnu} baseline across all four regions of interest, assessing both pixel-level segmentation and structured clinical feature prediction. Second, we designed a retrospective multi-reader, multi-case study involving radiologists of varying expertise to assess the association of OncoVision with diagnostic confidence, reading time, and segmentation accuracy, providing preliminary evidence of its utility as a decision-support tool. Third, we evaluated deployment feasibility by operationalizing OncoVision as a secure, user-friendly web application, now deployed at a partner hospital, capable of generating structured, interpretable reports in real-time.

The core contributions of this work are as follows:
\begin{itemize}
    \item Integrated multimodal pipeline: OncoVision jointly segments four regions of interest and predicts ten structured clinical features, including BI-RADS category, within a single integrated framework.
    
    \item Late-fusion evaluation: We propose and compare Independent and Dependent late-fusion variants that integrate imaging, clinical, and radiomic features during training, with the Dependent variant achieving higher accuracy on most tasks.
    
    \item Reader-assistance evaluation: In a paired reader-assistance evaluation, OncoVision was associated with higher diagnostic confidence for junior and senior radiologists and reduced reading time by up to 61\%.
    
    \item Deployable web application: OncoVision is deployed as a secure, Dockerized web platform that generates structured PDF reports with segmentation overlays, confidence scores, and attention-weighted visualizations.
\end{itemize}

By focusing on clinical utility and retrospective evaluation, this work advances AI-assisted mammographic interpretation, one that prioritizes interpretable and accessible diagnostic support over isolated technical benchmarks.

%%%%%%%%%% Figure 6: Integrated Workflow and Architecture %%%%%%%%%%
\begin{figure*}[t]
    \centering
    \includegraphics[width=\linewidth]{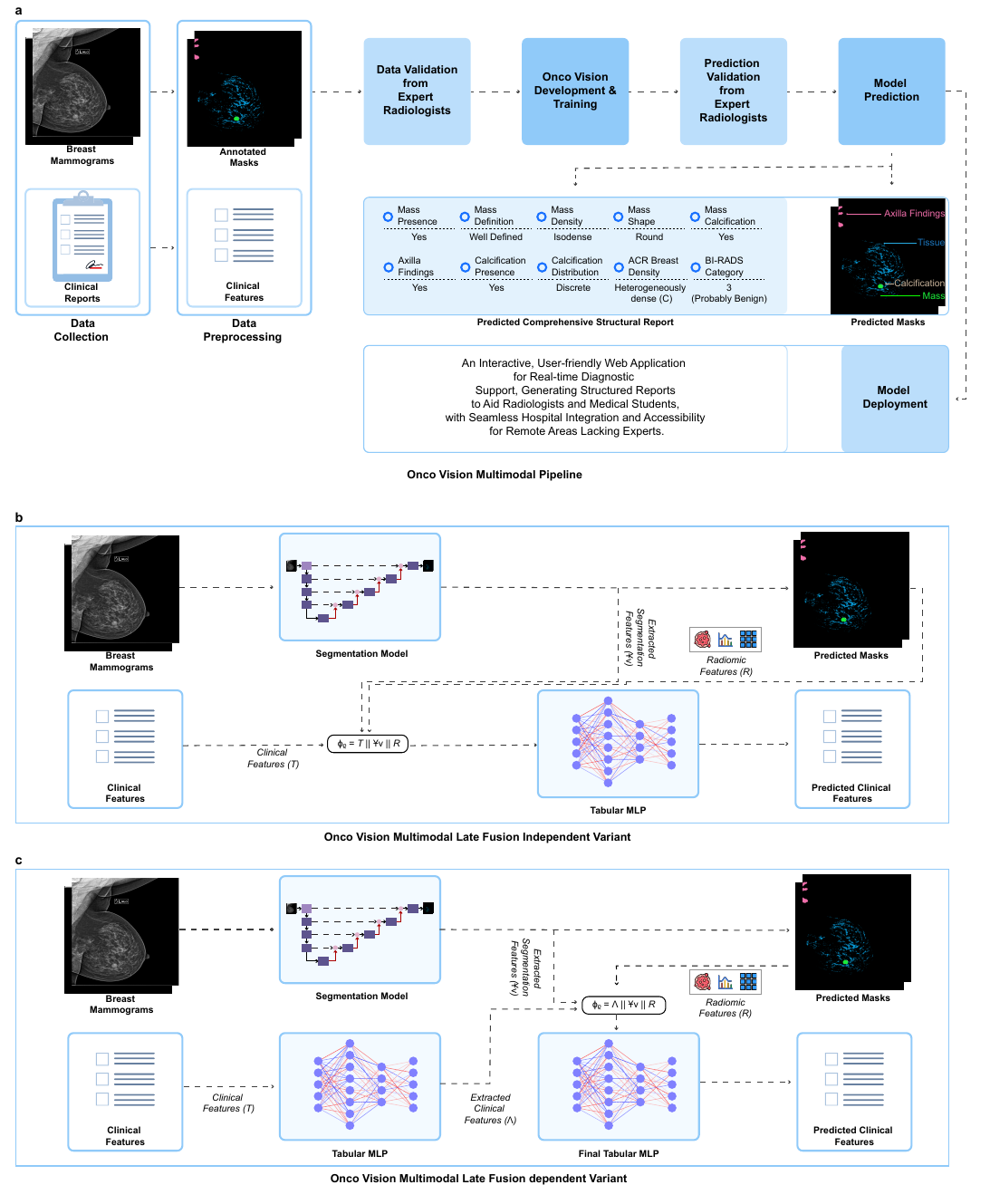}
    \caption{Integrated workflow and multimodal architecture of OncoVision. (a) Integrated workflow from data acquisition and preprocessing to model training, evaluation, and web deployment. (b) Independent variant: raw clinical features and radiomic descriptors are concatenated with imaging-derived bottleneck features during training. (c) Dependent variant: clinical features are processed through a dedicated encoder before fusion with imaging and radiomic features. At inference, externally supplied clinical features are omitted, while radiomic features are generated internally from the predicted segmentation masks.}
    \label{fig6}
\end{figure*}

\section{Related Work}\label{sec:related}

\subsection{Segmentation of Mammographic Regions of Interest}

Automated segmentation in mammography has largely targeted single regions of interest (ROIs) in isolation. Mass segmentation is the most studied task, with U-Net variants~\citep{ronneberger2015u, zhou2018unetplusplus, zhang2021unetpp} achieving IoU scores of 0.85--0.91 on benchmark datasets~\citep{zhang2021unetpp, singh2020breast}. Calcification detection has been explored for early DCIS identification, with reported AUCs approaching 0.95~\citep{samala2018deep, tong2024novel}. Breast tissue segmentation, motivated by density estimation, has achieved IoU around 0.78 via weight-adaptive multitask learning~\citep{gudhe2022area_based_density}. Axillary findings remain underexplored, with existing work limited to classification or detection of lymphadenopathy or fat-infiltrated nodes~\citep{song2023automated, tan2025mammography}, and no prior pixel-wise segmentation of axillary structures in mammograms.

Existing studies have not reported a unified framework that jointly segments all four ROIs while predicting a structured panel of clinical features. To the best of our knowledge, OncoVision is the first to combine these tasks in a single integrated pipeline, addressing a gap left by prior work that has focused on isolated tasks such as mass segmentation~\citep{zhang2021unetpp}, calcification detection~\citep{samala2018deep}, breast density estimation~\citep{gudhe2022area_based_density}, or BI-RADS categorization~\citep{tan2025mammography}, with axillary findings remaining particularly underexplored in segmentation~\citep{song2023automated}. OncoVision unifies masses, calcifications, axillary findings, and breast tissue in a shared attention-gated backbone, enabling contextual reasoning and supporting downstream volumetric and radiomic assessments.

\subsection{Structured Clinical Feature Prediction}

AI has been applied to predict individual clinical features from mammograms, including mass shape~\citep{singh2020breast}, calcification presence~\citep{tong2024novel}, breast density~\citep{gudhe2022area_based_density}, and BI-RADS category~\citep{tan2025mammography}. However, most systems predict only one or a few features and do not provide a structured diagnostic report~\citep{li2025birads}. Axillary findings, in particular, have been analyzed mostly for binary classification rather than integrated into multi-feature prediction~\citep{lundstrom2025lymph}. OncoVision departs from this fragmented approach by predicting ten clinically relevant features within a single pipeline, aligning with structured reporting standards and providing a diagnostic profile rather than isolated predictions.

\subsection{Multimodal Fusion and Clinical Deployment}

Multimodal fusion, combining imaging with clinical or tabular data, has shown promise in oncology, but most prior mammography work has focused on architectural novelty rather than clinical validation~\citep{multimodal_fusion_breast_2025}. Late fusion strategies, which integrate modality-specific representations after independent processing, offer a balance between predictive performance and robustness to missing data~\citep{multimodal_ai_scoping_2025}. OncoVision extends this approach with two late-fusion variants: an \textit{Independent} variant that concatenates raw clinical features and radiomic descriptors with imaging-derived representations, and a \textit{Dependent} variant that models clinical features through a dedicated encoder before fusion. Radiomic features, though widely used in oncology, remain underexplored as a complementary modality in integrated mammographic pipelines. From a deployment perspective, few AI systems for mammography have been operationalized as complete clinical tools; most remain research prototypes evaluated retrospectively~\citep{goh2025challenges}. OncoVision is implemented as a secure, Dockerized web application that generates structured PDF reports with segmentation overlays, confidence scores, and Grad-CAM visualizations, distinguishing it from prior proof-of-concept work.

\section{Methods}\label{sec4}

\subsection{Data}\label{subsec:data}

\subsubsection{Data Collection, Annotation, and Preprocessing}

We collected 1,725 mammograms from 500 patients at two institutions in Bangladesh: Bangladesh Specialized Hospital Limited (urban, Dhaka) and Khwaja Yunus Ali Medical College \& Hospital (rural, Rajshahi). Images were acquired in cranio-caudal and mediolateral oblique views. Expert radiologists excluded images with artifacts, poor contrast, or improper positioning, and clinical data were cross-referenced with electronic medical records. The study was approved by both hospitals' Institutional Review Boards, and data were de-identified.

To prevent leakage, all views from a patient were assigned to one split, with no patient appearing in more than one cohort. This patient-level partitioning is important in mammography, where multiple views of the same breast share characteristics that could inflate performance if distributed across splits. The dataset was partitioned at the patient level into training and test cohorts, with no patient appearing in more than one cohort. The test cohort comprised 100 patients, corresponding to 754 mammograms, and the remaining 400 patients formed the training cohort.

Segmentation masks for four regions of interest (masses, calcifications, axillary findings, and breast tissue) were created with pixel-level precision following American College of Radiology guidelines. Clinical features from structured reports included ten categorical variables: mass presence, definition, density, shape, calcification, axillary findings, calcification presence and distribution, ACR breast density, and BI-RADS category. Annotations were validated by two expert radiologists through consensus meetings, with high inter-annotator agreement (average Cohen's $\kappa = 0.97 \pm 0.01$). Challenging cases (isodense masses, faint microcalcifications below $0.1\%$ of image area, and ambiguous axillary findings) were refined under radiologist guidance.

Image preprocessing involved z-score normalization of pixel intensities, and masks were label-encoded and one-hot encoded for multi-class segmentation. To address severe class imbalance in clinically critical categories (e.g., BI-RADS 4: 5.7\%, BI-RADS 5: 1.0\%, spiculated masses: 0.5\%), targeted augmentation (rotation, flipping, intensity scaling, and controlled noise injection) was applied at the patient level to the full dataset, expanding it from 1,725 to 7,535 mammograms. The augmented dataset was then partitioned at the patient level into training and test cohorts, ensuring that no patient contributed views to both cohorts. All augmented views from a given patient were assigned to the same cohort as the original views. A class-aware oversampling strategy assigned augmented samples in inverse proportion to training-set frequency, so rare but clinically critical classes received more augmentation. Augmentations were applied consistently across views and ROIs. This expanded the training set 4.4-fold and balanced class representation; a control experiment without augmentation showed reduced performance on rare classes, confirming that gains reflect improved generalization rather than synthetic duplication. Continuous features with limited predictive power (e.g., patient age, mass size) were excluded from model input; mass size was instead derived post hoc from predicted segmentation masks.

\subsection{Model Architecture}\label{subsec:architecture}

OncoVision is a privileged-information training framework that uses mammographic imaging and internally derived radiomic descriptors at inference, while additionally using tabular clinical data during training. It is built on a unified attention-gated encoder--decoder backbone~\citep{gupta2026attention} that simultaneously segments four anatomical regions of interest (ROIs), masses, calcifications, axillary findings, and breast tissue, and predicts ten structured clinical features (Fig.~\ref{fig6}).

\subsubsection{Segmentation Backbone}

The segmentation backbone is a modified U-Net~\citep{ronneberger2015u} with encoder--decoder symmetry and integrated attention gates, designed to delineate anatomical and pathological regions with high boundary precision (Fig.~\ref{edfig4}a). The contracting path extracts hierarchical spatial features through sequential convolutional blocks, each comprising two $3 \times 3$ convolutions with Glorot uniform initialization, Leaky ReLU activation ($\alpha = 0.1$), batch normalization, and dropout. Filter counts increase progressively ($16 \to 32 \to 64 \to 128 \to 256$), while max-pooling reduces spatial resolution.

The bottleneck layer employs 512 filters to capture high-level semantic representations. The expansive path recovers spatial resolution via transposed convolutions, with attention gates at each upsampling step refining feature maps by suppressing irrelevant background. The refined attention map is given by $\tau_x = x \cdot \upsilon(\psi_f)$, where $x$ denotes the skip-connection feature map, $g$ the decoder gating signal, $\upsilon(\cdot)$ the sigmoid activation, and $\psi_f$ the attention coefficient computed from the combined transformed features of $x$ and $g$. This gating mechanism is particularly valuable for mammography, where low-contrast microcalcifications and isodense masses are easily obscured by surrounding fibroglandular tissue. The output layer applies a $1 \times 1$ convolution followed by softmax to produce a pixel-wise class probability map $\mathbf{P} \in \mathbb{R}^{H \times W \times \omega}$, where $\omega$ is the total number of segmentation classes (background, mass, calcification, axilla findings, breast tissue).

\subsubsection{Radiomic Feature Extraction}

After segmentation, radiomic descriptors are extracted from the predicted masks to provide explicit geometric, intensity, and texture information that complements the learned CNN representations during training. For each ROI, we compute shape descriptors (sphericity, perimeter-to-area ratio, eccentricity, compactness), first-order intensity statistics (mean, variance, skewness, kurtosis, entropy), and texture features derived from gray-level co-occurrence and run-length matrices. These descriptors are concatenated into a radiomic feature vector $R \in \mathbb{R}^{d_R}$, which is normalized and appended to the fused representation. Because radiomic features are computed from the model-predicted masks, they are derived internally from the mammogram and remain available at inference without requiring any externally supplied tabular data.

\subsubsection{Late Fusion Variants for Clinical Feature Prediction}

\begin{figure*}[t]
    \centering
    \includegraphics[width=\linewidth]{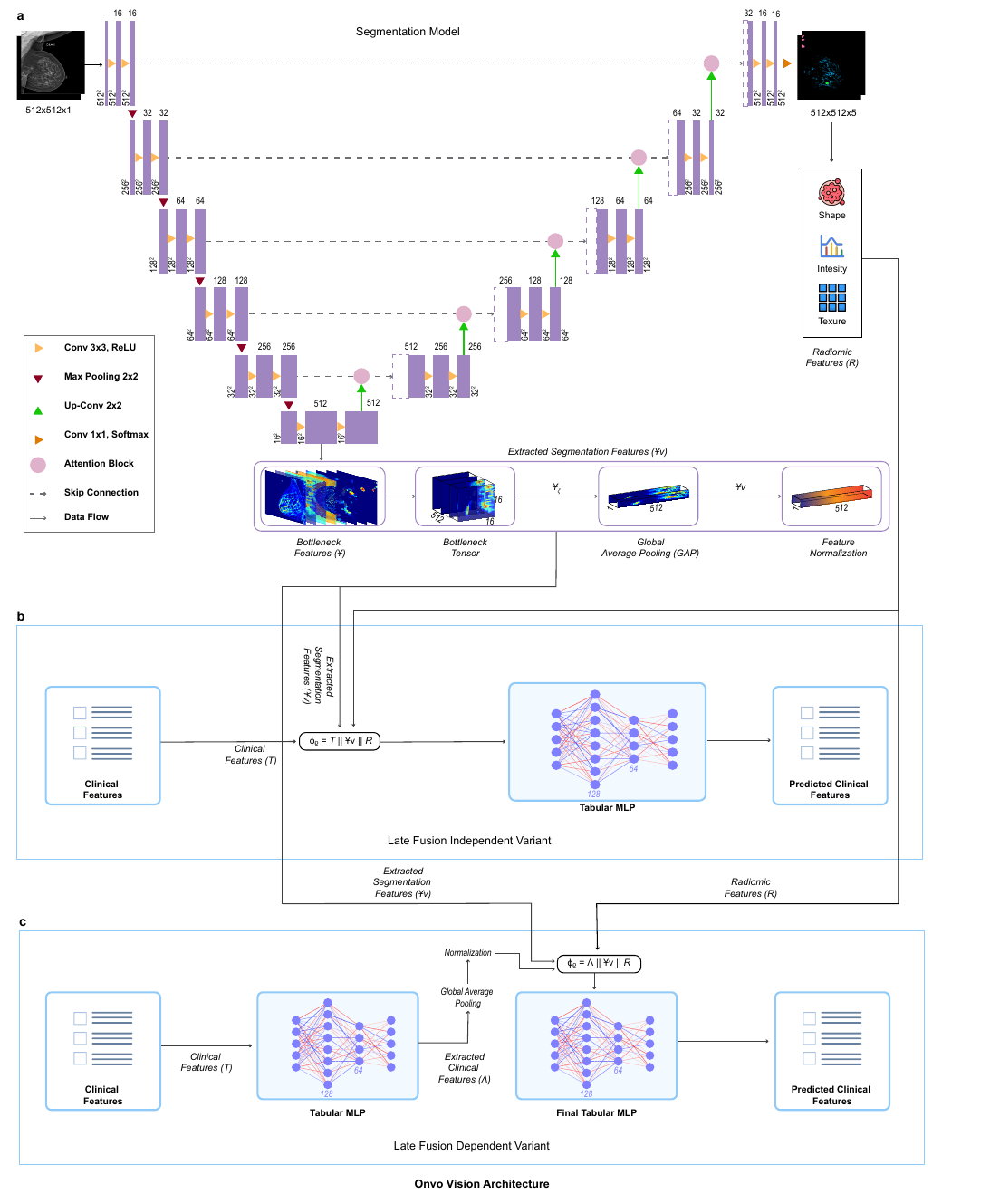}
    \caption{Detailed architecture of OncoVision. (a) Attention-gated U-Net segmentation backbone for four ROIs. (b) Independent variant: raw clinical features ($T$) and radiomic features ($R$) concatenated with imaging embeddings during training. (c) Dependent variant: clinical features processed through a dedicated encoder before fusion with imaging and radiomic features. At inference, externally supplied clinical features are omitted, while radiomic features are generated internally from the predicted segmentation masks and concatenated with the imaging representation before the corresponding inference head.}
    \label{edfig4}
\end{figure*}

OncoVision uses a privileged-information training strategy in which externally supplied clinical features are available during training but are not required at inference. Both variants share the same segmentation backbone and use radiomic descriptors derived from the predicted segmentation masks. Thus, at inference, the mammogram itself provides both the learned imaging representation and the radiomic descriptors through the predicted masks (Fig.~\ref{edfig4}b--c).

Independent variant. This variant concatenates the raw clinical feature vector with the imaging-derived bottleneck representation and radiomic descriptors during training. For an input mammogram $\iota$, the segmentation backbone extracts a bottleneck feature map $\Upsilon = \varsigma(\iota) \in \mathbb{R}^{H \times W \times C}$, where $H$, $W$, and $C$ denote height, width, and number of channels, respectively. Global average pooling compresses this feature map into a vector $\Upsilon_{\zeta} \in \mathbb{R}^{C}$, which is normalized to zero mean and unit variance to obtain $\Upsilon_{\nu}$. During training, the raw clinical feature vector $T \in \mathbb{R}^{d_T}$, normalized imaging representation $\Upsilon_{\nu}$, and radiomic vector $R \in \mathbb{R}^{d_R}$ are concatenated to form
\begin{equation}
    \Phi = T \,\mathbin{\|}\, \Upsilon_{\nu} \,\mathbin{\|}\, R.
\end{equation}
The fused representation is processed by a training head $\Xi$ to predict the ten clinical features, $\hat{Y}_{\mathrm{train}} = \Xi(\Phi)$. Because $T$ is unavailable at inference, a separate inference head $\Xi'$ is trained on the image-derived representation $\Phi' = \Upsilon_{\nu} \,\mathbin{\|}\, R$ and produces $\hat{Y}_{\mathrm{inf}} = \Xi'(\Phi')$. The two heads therefore differ in input dimensionality, while sharing the same image-derived representations.

Dependent variant. This variant first encodes the clinical feature vector through a dedicated tabular encoder. The tabular feature extractor $E_T(T)$ contains two hidden layers (128--64 units) and produces a compact latent representation $\Lambda = E_T(T)$, which is normalized before fusion. During training, the tabular latent representation, normalized imaging representation, and radiomic vector are concatenated:
\begin{equation}
    \Phi = \Lambda \,\mathbin{\|}\, \Upsilon_{\nu} \,\mathbin{\|}\, R.
\end{equation}
The resulting representation is processed by a three-layer MLP $\Xi_{\rho}$ (512--256--128 units) to predict the ten clinical features, $\hat{Y}_{\mathrm{train}} = \Xi_{\rho}(\Phi)$. Since $\Lambda$ requires the externally supplied clinical vector $T$, it is unavailable at inference. A separate inference head $\Xi_{\rho}'$, with input width $\dim(\Upsilon_{\nu})+\dim(R)$, is therefore trained on the reduced image-derived representation $\Phi' = \Upsilon_{\nu} \,\mathbin{\|}\, R$ and produces $\hat{Y}_{\mathrm{inf}} = \Xi_{\rho}'(\Phi')$. No externally supplied clinical features are required by either inference head.

At inference, the mammogram is passed through the segmentation backbone to obtain predicted masks $\hat{M}$. Radiomic descriptors $R$ are then computed from these predicted masks, while the same mammogram is processed by the shared backbone to obtain $\Upsilon_{\nu}$. The resulting image-derived representation $\Upsilon_{\nu}\Vert R$ is passed to the corresponding inference head $\Xi'$ or $\Xi_{\rho}'$ to predict the ten clinical features. Thus, clinical features are privileged training information, whereas radiomic descriptors are internally generated image-derived features available at both training and inference.

\subsubsection{Loss Function}

The model is trained jointly with a composite loss combining segmentation and clinical prediction objectives:
\begin{equation}
    L_{\mathbf{O}} = \lambda_1 (L_{\mathbf{D}} + L_{\mathbf{F}}) + \lambda_2 L_{\Gamma},
\end{equation}
where $L_{\mathbf{D}}$ is the Dice loss for segmentation overlap, $L_{\mathbf{F}}$ is the focal loss for class imbalance, and $L_{\Gamma}$ is the categorical cross-entropy for clinical feature prediction. The Dice loss maximizes overlap between predicted masks $\hat{M}$ and ground-truth masks $M_{\text{gt}}$, $L_{\mathbf{D}} = 1 - (2|\hat{M} \cap M_{\text{gt}}| + \epsilon) / (|\hat{M}| + |M_{\text{gt}}| + \epsilon)$, with $\epsilon = 10^{-5}$ for numerical stability. The focal loss, $L_{\mathbf{F}} = -\alpha \sum_i (1 - p_i)^\gamma \log(p_i)$, uses $\alpha = 0.5$ and $\gamma = 2$, where $p_i$ is the predicted probability of the true class for pixel $i$, to prioritize hard-to-classify pixels such as faint calcifications. The classification loss is standard categorical cross-entropy, $L_{\Gamma} = -\sum_i Y_i^{\mathrm{gt}} \log(\hat{Y}_i)$, where $Y_i^{\mathrm{gt}}$ is the ground-truth clinical label for class $i$ and $\hat{Y}_i$ is the predicted probability. The classification loss is applied jointly to the outputs of the training and inference heads of both variants ($\Xi$ and $\Xi'$, $\Xi_\rho$ and $\Xi_\rho'$), so that the deployed inference heads learn from the same supervision as their training counterparts. Hyperparameters $\lambda_1 = 0.7$ and $\lambda_2 = 0.3$ were empirically selected to balance segmentation accuracy and clinical prediction reliability.

%%%%% Figure 7: Detailed Architecture %%%%%

\subsection{Ablation Study}\label{sec:ablation}

To evaluate the contribution of each architectural component to clinical report prediction, we conducted a cumulative ablation study using the same patient-disjoint training, validation, and test partitions as the main experiments. The baseline configuration consists of nnU-Net for segmentation with an MLP head attached to the bottleneck features for clinical prediction, without any late fusion. Each row in Table~\ref{tab:abl:roadmap} adds one component to the configuration above, isolating the incremental benefit for clinical prediction. All ablation configurations were trained on the training partition with k-fold cross-validation for hyperparameter selection and evaluated on the held-out test partition. For consistency with the main results, ablation performance is reported on the held-out test partition; however, the final OncoVision architecture was fixed prior to test-set ablation evaluation and was not modified based on these results. The reader study and primary results in Section~\ref{sec:results} use this fixed configuration. Clinical accuracy is the mean accuracy across the ten structured features (macro-averaged F1 for multi-class tasks), and BI-RADS Macro F1 corresponds to the ordinal BI-RADS category prediction.

%%%%%%%%%% Table 3: Clinical Report Prediction Roadmap %%%%%%%%%%
\begin{table*}[t]
\centering
\caption{Clinical report prediction roadmap. Each row adds one component to the configuration above. Clinical accuracy is the mean accuracy across the ten structured features (macro-averaged F1 for multi-class tasks). BI-RADS Macro F1 corresponds to the ordinal BI-RADS category prediction. Mass Dice and HD remain constant across rows sharing the same segmentation backbone. Highlighted cells mark the best value per metric.}
\label{tab:abl:roadmap}
\small
\setlength{\tabcolsep}{5pt}
\begin{tabular}{llcccc}
\toprule
\# & Configuration & Mass Dice$\uparrow$ & Mass HD$\downarrow$ & BI-RADS Macro F1$\uparrow$ & Clin. Acc$\uparrow$ \\
\midrule
0 & nnU-Net + MLP (no fusion) & 0.9353 & 10.53 & 0.5812 & 0.7184 \\
1 & nnU-Net + Independent & 0.9353 & 10.53 & 0.6724 & 0.8031 \\
2 & nnU-Net + Dependent & 0.9353 & 10.53 & 0.6887 & 0.8172 \\
3 & OncoVision + Dependent & 0.9521 & 7.82 & 0.8104 & 0.8823 \\
4 & OncoVision + Independent & 0.9521 & 7.82 & 0.7721 & 0.8604 \\
5 & OncoVision + Dependent + attention & 0.9521 & 7.82 & 0.8241 & 0.8912 \\
6 & \textbf{OncoVision + Dependent + attention + radiomic concat.} & \cellcolor{bestblue}$\bm{0.9521}$ & \cellcolor{bestblue}$\bm{7.82}$ & \cellcolor{bestblue}$\bm{0.8552}$ & \cellcolor{bestblue}$\bm{0.9057}$ \\
\bottomrule
\end{tabular}
\end{table*}

The roadmap shows that clinical prediction accuracy improves from 0.7184 (nnU-Net with MLP head, no fusion) to 0.9057 (full OncoVision with Dependent fusion, attention gates, and radiomic concatenation) a gain of 18.7 percentage points. BI-RADS Macro F1 improves from 0.5812 to 0.8552. Segmentation performance also improves, with mass Dice increasing from 0.9353 to 0.9521 and HD decreasing from 10.53 to 7.82 pixels. At the final configuration, the false positive rate and false negative rate across all clinical features are 0.0321 and 0.0541, respectively.

\subsubsection{Effect of Segmentation Backbone}

The OncoVision attention-gated backbone consistently outperforms nnU-Net for downstream clinical prediction (Table~\ref{tab:ablation_backbone}). With Dependent fusion, OncoVision improves clinical accuracy from 0.8172 to 0.8823 (+6.5 points) and BI-RADS Macro F1 from 0.6887 to 0.8104 (+12.2 points). Segmentation Dice also improves from 0.9353 to 0.9521, and HD decreases from 10.53 to 7.82 pixels.

\begin{table}[!htb]
\centering
\footnotesize
\setlength{\tabcolsep}{3pt}
\resizebox{\columnwidth}{!}{%
\begin{tabular}{llcc}
\toprule
\textbf{Backbone} & \textbf{Fusion} & \textbf{BI-RADS Macro F1$\uparrow$} & \textbf{Clin. Acc$\uparrow$} \\
\midrule
nnU-Net & Independent & 0.6724 & 0.8031 \\
nnU-Net & Dependent & 0.6887 & 0.8172 \\
OncoVision & Independent & 0.7721 & 0.8604 \\
OncoVision & Dependent & \textbf{0.8104} & \textbf{0.8823} \\
\bottomrule
\end{tabular}%
}
\caption{\textbf{Ablation on backbone and fusion strategy.} OncoVision with Dependent fusion achieves the best clinical prediction performance. Bold indicates the best-performing configuration.}
\label{tab:ablation_backbone}
\end{table}

\subsubsection{Effect of Fusion Strategy}

Comparing Independent and Dependent fusion on the OncoVision backbone, Dependent fusion achieves higher clinical accuracy (0.8823 vs. 0.8604) and higher BI-RADS Macro F1 (0.8104 vs. 0.7721). This confirms the Dependent variant as the preferred configuration. Because the Independent variant performed poorly, we did not extend it with attention gates or radiomic fusion.

\subsubsection{Effect of Attention Gates}

Adding attention gates to the Dependent variant improved clinical accuracy from 0.8823 to 0.8912 (+0.9 points) and BI-RADS Macro F1 from 0.8104 to 0.8241 (+1.4 points). Segmentation performance also improved, with calcification Dice increasing from 0.7985 to 0.8413.

\begin{table}[!htb]
\centering
\footnotesize
\setlength{\tabcolsep}{3pt}
\resizebox{\columnwidth}{!}{%
\begin{tabular}{lcc}
\toprule
\textbf{Configuration} & \textbf{BI-RADS Macro F1$\uparrow$} & \textbf{Clin. Acc$\uparrow$} \\
\midrule
OncoVision + Dependent & 0.8104 & 0.8823 \\
+ Attention gates & \textbf{0.8241} & \textbf{0.8912} \\
\bottomrule
\end{tabular}%
}
\caption{\textbf{Ablation on attention gates.} Adding attention gates improves clinical prediction and BI-RADS Macro F1. Bold indicates the best-performing configuration.}
\label{tab:ablation_attention}
\end{table}

\subsubsection{Effect of Radiomic Integration}

Adding radiomic features extracted from predicted masks via concatenation to the attention-augmented Dependent variant improved clinical accuracy from 0.8912 to 0.9057 (+1.5 points) and BI-RADS Macro F1 from 0.8241 to 0.8552 (+3.1 points). The largest gains were observed for BI-RADS Category, Mass Definition, and Mass Calcification, suggesting that explicit shape and texture descriptors complement the learned CNN representations for morphologically complex and high-risk categories. Radiomic concatenation was associated with lower false-positive and false-negative rates across the evaluated clinical features.

\begin{table}[!htb]
\centering
\footnotesize
\setlength{\tabcolsep}{3pt}
\resizebox{\columnwidth}{!}{%
\begin{tabular}{lcc}
\toprule
\textbf{Radiomic Strategy} & \textbf{BI-RADS Macro F1$\uparrow$} & \textbf{Clin. Acc$\uparrow$} \\
\midrule
No radiomics & 0.8241 & 0.8912 \\
\textbf{Concatenation} & \cellcolor{bestblue}$\bm{0.8552}$ & \cellcolor{bestblue}$\bm{0.9057}$ \\
\bottomrule
\end{tabular}%
}
\caption{\textbf{Ablation on radiomic integration.} Concatenation of radiomic features improves clinical prediction and BI-RADS Macro F1. Bold indicates the best-performing configuration.}
\label{tab:ablation_radiomics}
\end{table}

\subsubsection{Effect of Multi-ROI Joint Segmentation}

To assess whether joint segmentation of all four ROIs contributes to clinical prediction, we removed individual ROIs from the segmentation backbone and measured the impact on clinical prediction accuracy. Removing mass segmentation caused the largest drop ($-4.16$ points), followed by calcification ($-3.07$ points), confirming these features are most informative for downstream clinical reasoning. Breast tissue removal had the smallest impact ($-0.80$ points), consistent with its contextual rather than diagnostic role. Notably, axilla removal caused a disproportionate drop in axillary findings prediction, demonstrating that explicit axilla segmentation provides critical spatial context for staging-related predictions.

%%%%%%%%%% Table 1: Segmentation Performance (Spanning Two Columns) %%%%%%%%%%
\begin{table*}[t]
\centering
\scriptsize
\setlength{\tabcolsep}{4pt}
\begin{tabular}{lccccccccccc}
\toprule
\textbf{\multirow{2}{*}{ROI}} & \textbf{\multirow{2}{*}{$\bm{\uparrow}$IoU}} & \textbf{\multirow{2}{*}{$\bm{\uparrow}$Dice}} & \textbf{\multirow{2}{*}{$\bm{\uparrow}$Precision}} & \textbf{\multirow{2}{*}{$\bm{\uparrow}$Sensitivity}} & \textbf{\multirow{2}{*}{\makecell[c]{$\bm{\uparrow}$F1\\Score}}} & \textbf{\multirow{2}{*}{$\bm{\uparrow}$Specificity}} & \textbf{\multirow{2}{*}{\makecell[c]{$\bm{\downarrow}$HD\\(px)}}} & \textbf{\multirow{2}{*}{\makecell[c]{$\bm{\downarrow}$ASD\\(px)}}} & \textbf{\multirow{2}{*}{\makecell[c]{$\bm{\uparrow}$Boundary\\IoU}}} & \textbf{\multirow{2}{*}{\makecell[c]{RVD\\(mean ± std)}}} & \textbf{\multirow{2}{*}{\makecell[c]{RAVD\\(mean ± std)}}} \\ 
&&&&&&&&&&&\\
\toprule
\multicolumn{12}{c}{\textbf{nnU-Net}} \\ \midrule
 Mass & 0.8785 & 0.9353 & 0.9423 & 0.9313 & 0.9361 & 0.9794 & 10.5253 & 1.3192 & 0.2973 & $-0.02 \pm 0.16$ & $0.05 \pm 0.15$ \\ 
 Axilla Findings & 0.7413 & 0.8514 & 0.8728 & 0.8356 & 0.8531 & 0.9719 & 25.1785 & 4.1672 & 0.2992 & $-0.04 \pm 0.21$ & $0.09 \pm 0.18$ \\ 
 Breast Tissue & 0.6138 & 0.7607 & 0.7854 & 0.7386 & 0.7615 & 0.9457 & 55.2629 & 2.0713 & 0.3222 & $+0.01 \pm 0.42$ & $0.21 \pm 0.37$ \\ 
 Calcification & 0.6485 & 0.7868 & 0.8217 & 0.7561 & 0.7877 & 0.9811 & 95.0511 & 18.7892 & 0.2103 & $-0.22 \pm 0.39$ & $0.25 \pm 0.36$ \\  \midrule
\multicolumn{12}{c}{\textbf{OncoVision}} \\ \midrule
 Mass & \textbf{0.9125} & \textbf{0.9521} & \textbf{0.9622} & \textbf{0.9424} & \textbf{0.9522} & \textbf{0.9881} & \textbf{7.8171} & \textbf{1.0098} & \textbf{0.3151} & $-0.015 \pm 0.152$ & $0.041 \pm 0.147$ \\ 
 Axilla Findings & \textbf{0.7839} & \textbf{0.8747} & \textbf{0.9020} & \textbf{0.8491} & \textbf{0.8747} & \textbf{0.9752} & \textbf{20.6907} & \textbf{3.5071} & \textbf{0.3188} & $-0.035 \pm 0.204$ & $0.080 \pm 0.191$ \\ 
 Breast Tissue & \textbf{0.6565} & \textbf{0.7926} & \textbf{0.8145} & \textbf{0.7720} & \textbf{0.7927} & \textbf{0.9541} & \textbf{45.0871} & \textbf{1.5258} & \textbf{0.3525} & $+0.007 \pm 0.410$ & $0.190 \pm 0.364$ \\ 
 Calcification & \textbf{0.7351} & \textbf{0.8413} & \textbf{0.8923} & \textbf{0.7959} & \textbf{0.8413} & \textbf{0.9845} & \textbf{83.3120} & \textbf{15.5355} & \textbf{0.2361} & $-0.200 \pm 0.379$ & $0.236 \pm 0.357$ \\
 \bottomrule
\end{tabular}%
\caption{
\textbf{Segmentation performance comparison between nnU-Net and OncoVision across four regions of interest (ROIs).}
Metrics include IoU, Dice, Precision, Sensitivity, F1, Specificity, Hausdorff Distance (HD), Average Surface Distance (ASD), Boundary IoU, and Relative Volume Difference (RVD/RAVD, reported as mean~$\pm$~std). Arrows ($\bm{\uparrow}$/$\bm{\downarrow}$) indicate whether larger or smaller values are preferred. Bold indicates the best-performing model. OncoVision achieves consistent gains across all ROIs, with the largest improvements for calcification and mass segmentation.}
\label{tab1:segmentation_performance_metrics}
\end{table*}

\section{Results}\label{sec:results}

We evaluated OncoVision on a held-out test set of 754 mammograms from 100 patients not seen during training, assessing segmentation accuracy across four regions of interest (ROIs) and clinical feature prediction across ten structured diagnostic tasks.

\subsection{Segmentation Performance}

\begin{figure*}[t]
    \centering
    \includegraphics[width=\linewidth]{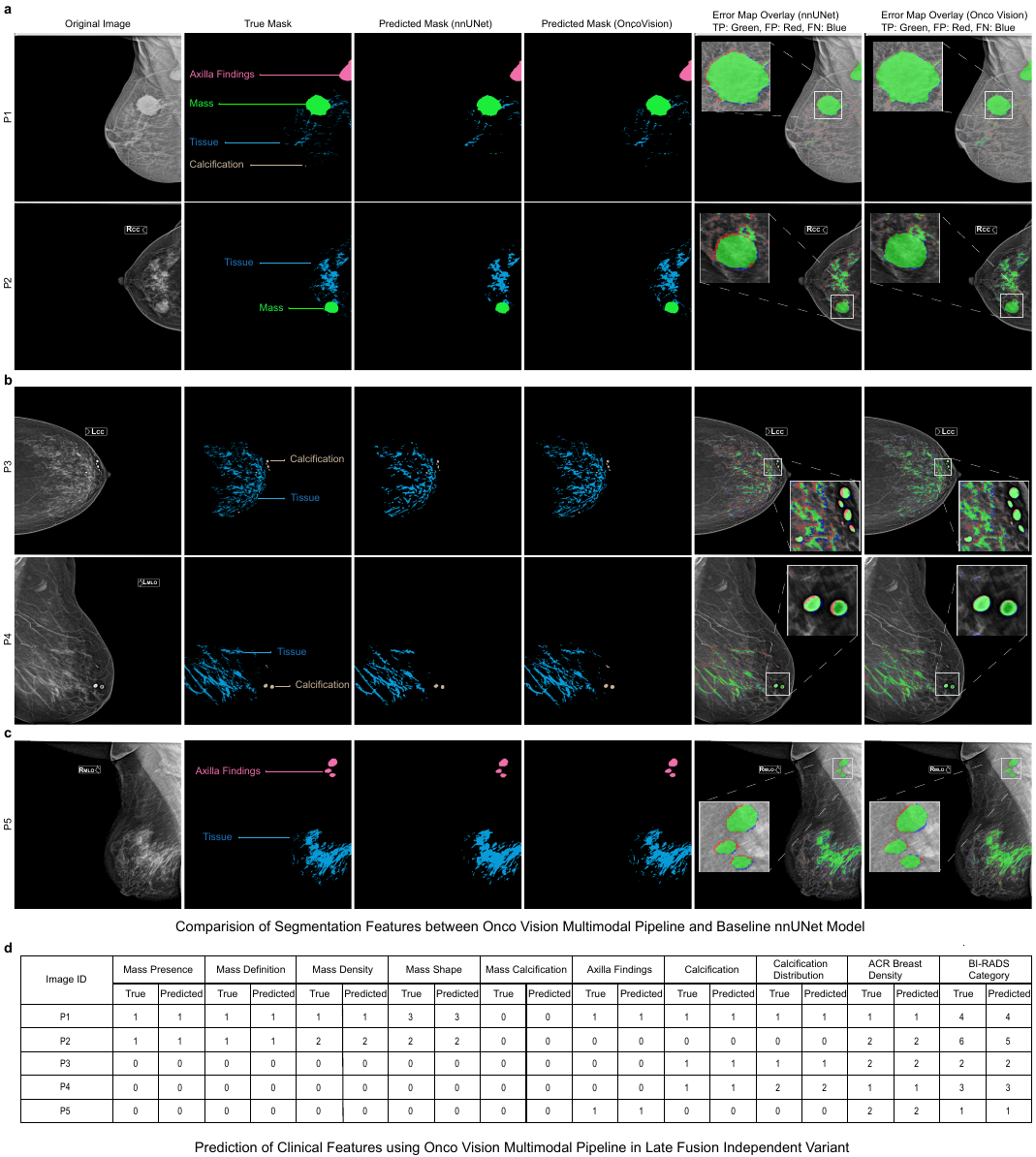}
    \caption{Comparative segmentation and diagnostic profiling using OncoVision. (a--c) Side-by-side segmentation comparison of nnU-Net and OncoVision across mass (a), calcification (b), and axilla findings with breast tissue (c). Panels show, from left to right: original mammogram, ground truth, nnU-Net prediction, OncoVision prediction, nnU-Net error map, and OncoVision error map (true positives in green, false positives in red, false negatives in blue). Zoomed insets (3--4$\times$) highlight improved boundary delineation and reduced false errors by OncoVision. (d) Clinical feature predictions by the Dependent variant for five patients. All predictions match ground truth except one BI-RADS downgrade (P2: 6 $\rightarrow$ 5). Data are from the test set of 754 mammograms (100 patients).}
    \label{fig1}
\end{figure*}

OncoVision outperformed nnU-Net across all four ROIs on every segmentation metric (Table~\ref{tab1:segmentation_performance_metrics}; Fig.~\ref{fig2}a). Mass segmentation reached an IoU of 0.9125 and DSC of 0.9522, compared with 0.8785 and 0.9353 for nnU-Net clinically relevant for surgical planning and biopsy guidance. Calcifications, the most challenging ROI due to their small size and low contrast, showed the largest improvement (IoU: 0.7351 vs. 0.6485; DSC: 0.8413 vs. 0.7868), reflecting improved segmentation of subtle, low-contrast calcification clusters. Axillary findings showed similar benefits (IoU: 0.7839 vs. 0.7413), while breast tissue showed the smallest relative improvement (IoU: 0.6565 vs. 0.6138), consistent with its visually simpler structure. Boundary metrics confirmed these gains, with lower Hausdorff Distance (7.82 vs. 10.53 pixels for mass) and Average Surface Distance (1.01 vs. 1.32 pixels). Visual error analysis (Fig.~\ref{fig1}a--c) supported these findings: OncoVision reduced false positives and negatives at mass margins, reduced false negatives for low-contrast calcification clusters, and localized axillary lymph nodes with fewer false positives.

\subsection{Clinical Feature Prediction}

OncoVision's Dependent variant predicted all ten clinical features, producing structured diagnostic reports that matched ground-truth labels (Fig.~\ref{fig1}d; Table~\ref{tab2}). For patients P1--P5, predictions matched ground truth across all features except one non-actionable BI-RADS downgrade (P2: 6 $\rightarrow$ 5).

The Dependent variant performed well on binary tasks (accuracy: 0.9610 for mass presence, 0.9801 for mass calcification; Table~\ref{tab2}). For multi-class tasks, performance varied by class complexity: mass definition achieved the highest macro F1 (0.8830), while BI-RADS category reached 0.8552, with lower recall observed for rare, morphologically complex categories such as spiculated masses and BI-RADS 5, reflecting both the inherent difficulty of these classes and their underrepresentation despite targeted augmentation.

ROC analysis indicated strong discriminative performance, with mean AUCs above 0.91 across all ten features. Performance was highest for mass calcification (AUC = 0.979), axilla findings (AUC = 0.978), and line/segmental calcification distribution (AUC = 0.991) a challenging, low-contrast pattern that is difficult to characterize reliably. For BI-RADS, high-risk categories achieved AUCs of 0.941 (category 4), 0.927 (category 5), and 0.917 (category 6), suggesting potential utility for risk stratification. However, per-class recall for rare, morphologically complex classes remains a limitation despite the improved overall accuracy.

\begin{figure*}[t]
    \centering
    \includegraphics[width=\linewidth]{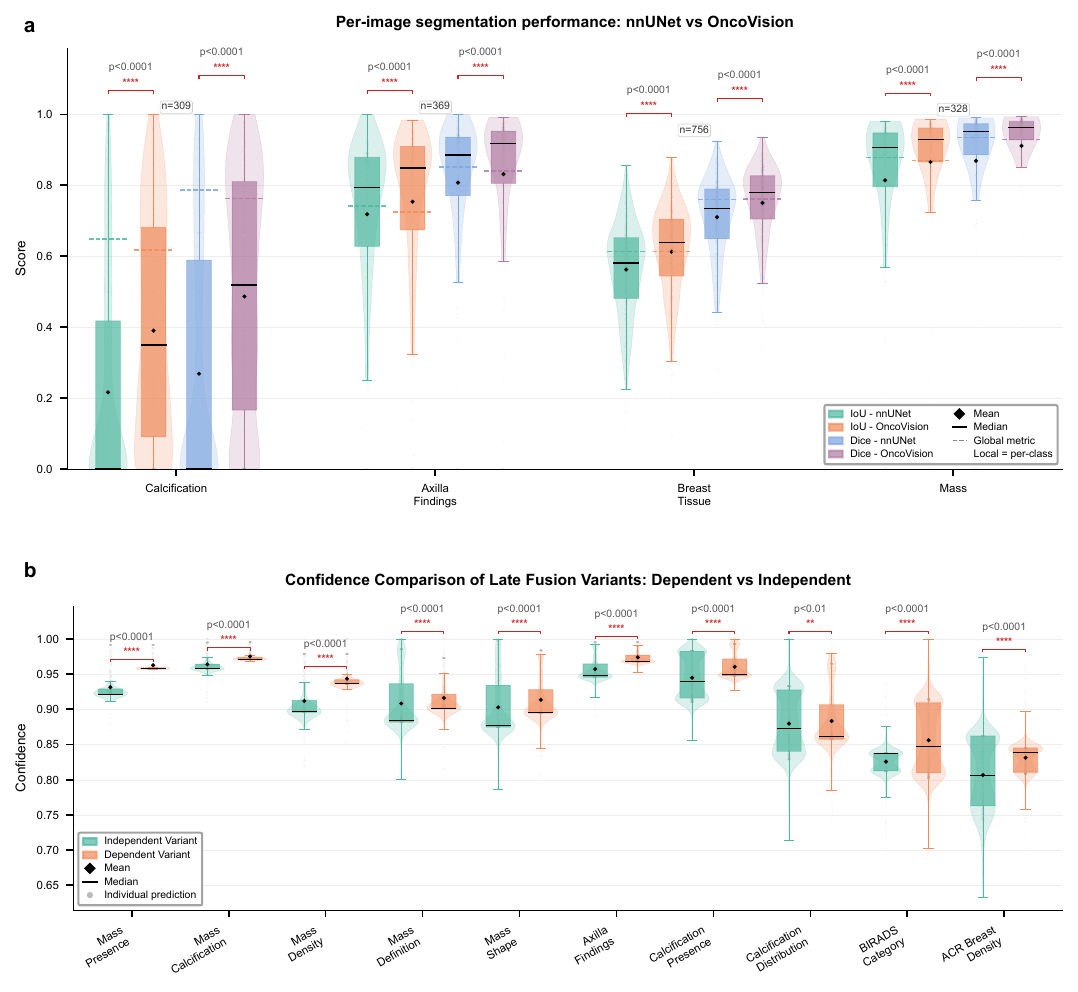}
    \caption{Comparative per-image segmentation and clinical feature confidence of nnU-Net and OncoVision. (a) Box plots of per-image IoU and DSC for calcification ($n = 309$), axilla findings ($n = 369$), breast tissue ($n = 754$), and mass ($n = 328$). Red dashed/solid lines: nnU-Net; blue dashed/solid lines: OncoVision. Wilcoxon signed-rank test; $^{***}P < 0.001$; $^{**}P < 0.01$; $^{*}P < 0.05$. Large effect sizes for calcification (DSC: Cohen's $d = -0.877$) and breast tissue (IoU: $d = -0.857$). (b) Box plots of model confidence across ten clinical features ($n = 754$ each) for the Dependent (green) and Independent (orange) variants after radiomic feature fusion. The Dependent variant showed significantly higher confidence across all ten features (Wilcoxon signed-rank test; $p < 0.0001$ for nine features; $p < 0.01$ for Calcification Distribution). Mean differences ranged from $0.0038$ (Calcification Distribution) to $0.0316$ (Mass Density), with large paired effect sizes for Mass Presence ($d = 1.76$), Mass Density ($d = 1.04$), and Axilla Findings ($d = 1.02$). These results show that radiomic feature fusion improved model confidence across both binary and multi-class clinical prediction tasks.}
    \label{fig2}
\end{figure*}

The Independent variant reported slightly lower accuracy for mass presence (0.9556 vs. 0.9610) but marginally higher accuracy for mass calcification (0.9829 vs. 0.9801), suggesting a more conservative prediction strategy. For ACR breast density, the Dependent variant achieved a higher macro F1 (0.8247 vs. 0.7893). Model confidence distributions (Fig.~\ref{fig2}b) showed the Dependent variant consistently more confident across all tasks, with the largest differences for BI-RADS category and ACR breast density.

\begin{table*}[t]
\centering
\footnotesize
\setlength{\tabcolsep}{6pt}
\begin{tabular}{llcccc}
\toprule
\textbf{Clinical Feature} & \textbf{Metric} & \textbf{Dependent} & \textbf{Independent} & \textbf{$\Delta$} & \textbf{Preferred} \\
\midrule
\multicolumn{6}{l}{\textit{\textbf{Binary Tasks (Yes/No)}}} \\
\midrule
Mass Presence & Accuracy & \textbf{0.9610} & 0.9556 & $+0.0054$ & Dependent \\
Mass Calcification & Accuracy & 0.9801 & \textbf{0.9829} & $-0.0028$ & Independent \\
Axillary Findings & Accuracy & \textbf{0.9537} & 0.9322 & $+0.0215$ & Dependent \\
Calcification Presence & Accuracy & \textbf{0.9356} & 0.9209 & $+0.0147$ & Dependent \\
\midrule
\multicolumn{6}{l}{\textit{\textbf{Multi-Class Tasks (Macro-Averaged F1)}}} \\
\midrule
Mass Definition & Macro F1 & \textbf{0.8830} & 0.7748 & $+0.1082$ & Dependent \\
Mass Density & Macro F1 & \textbf{0.8977} & 0.8648 & $+0.0329$ & Dependent \\
Mass Shape & Macro F1 & \textbf{0.9071} & 0.8797 & $+0.0274$ & Dependent \\
Calcification Distribution & Macro F1 & \textbf{0.8624} & 0.8111 & $+0.0513$ & Dependent \\
ACR Breast Density & Macro F1 & \textbf{0.8247} & 0.7893 & $+0.0354$ & Dependent \\
BI-RADS Category & Macro F1 & \textbf{0.8552} & 0.8061 & $+0.0491$ & Dependent \\
\bottomrule
\end{tabular}
\caption{\textbf{Clinical feature prediction performance of OncoVision's two late-fusion variants with radiomic features.} Binary tasks are reported as accuracy; multi-class tasks as macro-averaged F1. The Dependent variant outperforms the Independent variant on 9 of 10 features. Bold indicates the best-performing variant.}
\label{tab2}
\end{table*}

\subsubsection{Study Design and Participants}

Six board-certified radiologists were enrolled and stratified by experience: two juniors (1--5 years post-residency), two seniors (8--10 years), and two experts (15--25 years). A reference set of 100 retrospectively labeled mammographic cases was selected to represent a heterogeneous population, including benign and malignant lesions, high and low breast density, and simple and complex diagnostic scenarios. Radiologists independently evaluated each case in a blinded fashion and completed a structured report covering ten clinical items: mass presence, definition, density, shape, and calcification; axillary findings; calcification presence and distribution; ACR breast density; and BI-RADS category. For each feature, readers assigned a confidence score on a 0--1 scale. Reading time per case was recorded using preconfigured software timers.

%%%%%%%%%% Reader Study Figure %%%%%%%%%
\begin{figure*}[t]
    \centering
    \includegraphics[width=\linewidth]{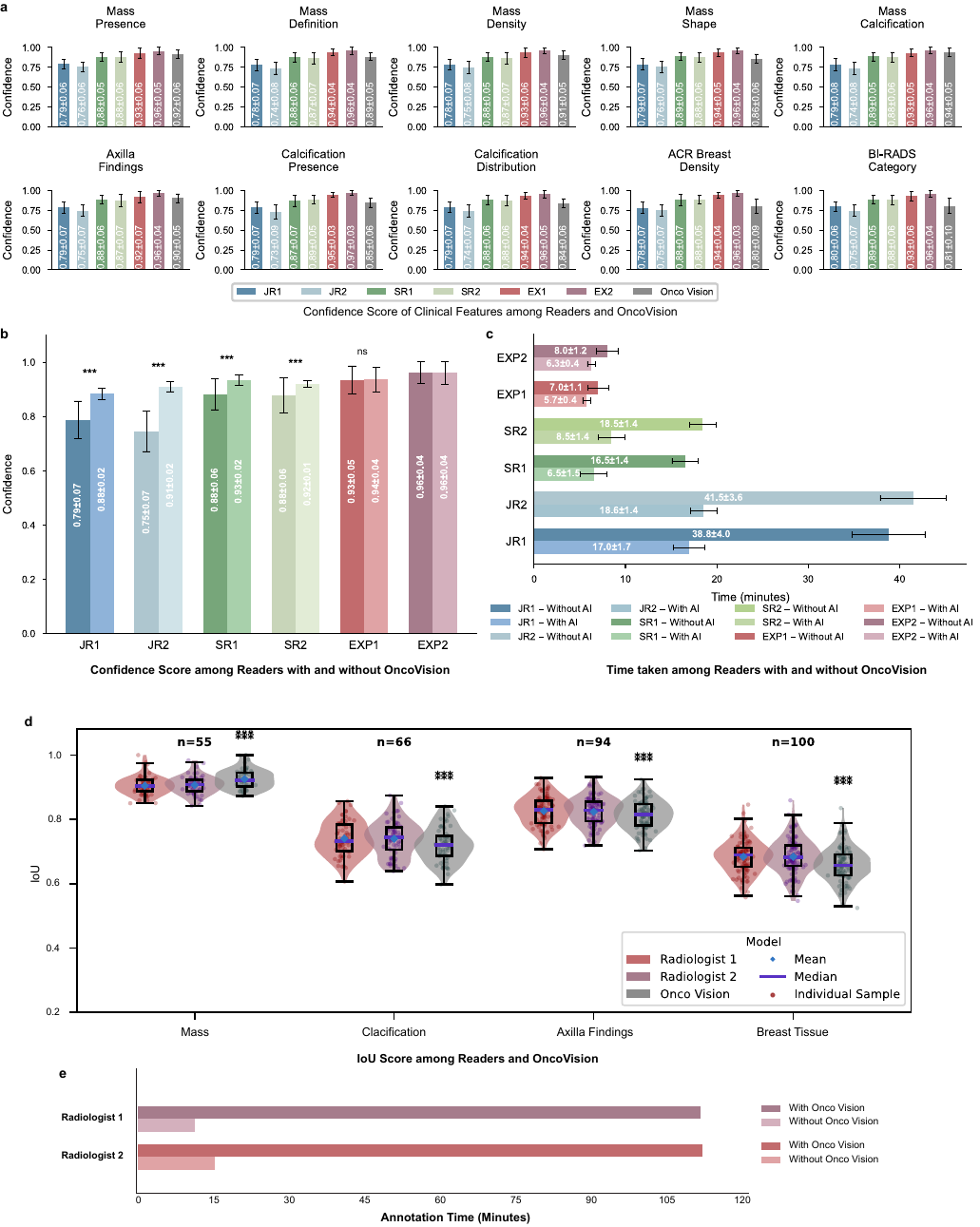}
    \caption{Paired reader-assistance evaluation of OncoVision as a decision-support tool. (a) Diagnostic confidence across ten clinical features for six radiologists (JR: junior; SR: senior; EXP: expert) and OncoVision. (b) Paired confidence with/without AI assistance. (c) Diagnostic time with/without AI. (d) Segmentation IoU vs. radiologists across four ROIs. (e) Batch segmentation time. Statistical details are provided in the Reader Study section.}
    \label{edfig3}
\end{figure*}

OncoVision processed the same cases, generating predictions with dual-confidence scores from both late-fusion variants. In a paired analysis, readers repeated assessments with OncoVision assistance, accessing its segmented outputs, predicted features, and visualizations via the web application. AI outputs were presented alongside images in a decision-support mode, and final interpretations remained at the reader's discretion. Re-reading the same cases introduces potential learning and recall effects, and a washout period was not enforced due to the retrospective design. Readers were blinded to their initial assessments and to ground-truth labels during the AI-assisted phase. The study should therefore be interpreted as a paired reader-assistance evaluation rather than a definitive clinical validation; prospective multi-center trials with case randomization and washout periods are needed to confirm diagnostic benefit. For segmentation evaluation, two additional radiologists (5--10 years' experience) manually annotated 100 complex cases featuring isodense masses, micro-discrete calcifications, and intricate breast tissue structures, and their annotations were compared against OncoVision's automated segmentations for four regions of interest (ROIs): masses, calcifications, axilla findings, and breast tissue.

Statistical analyses were performed in Python. Confidence scores and reading times were compared using Wilcoxon signed-rank tests (paired for with/without AI; unpaired for group comparisons), with one-tailed tests and Bonferroni correction where applicable ($\alpha = 0.05/8 = 0.00625$ for segmentation). Mixed-effects models estimated the interaction between reader expertise and AI assistance, treating reader and case as random effects. Effect sizes were quantified using Cohen's $d$ (small: $|d| < 0.5$; medium: 0.5--0.8; large: $> 0.8$). All confidence intervals are reported at the 95\% level. Intersection-over-Union (IoU) scores assessed segmentation agreement against ground truth.

\subsubsection{Diagnostic Confidence}

OncoVision showed confidence levels that varied by reader expertise across the ten clinical features (Fig.~\ref{edfig3}a). Mean confidence ($\pm$ s.d.) averaged over 100 cases was $0.871 \pm 0.077$ for OncoVision, compared with $0.767 \pm 0.071$ for juniors, $0.880 \pm 0.061$ for seniors, and $0.948 \pm 0.046$ for experts. Pairwise comparisons indicated higher confidence for OncoVision versus juniors in 8/10 features ($P < 0.001$, Cohen's $d = 0.856$--$2.289$), mixed results versus seniors (5/10 features significant; $P < 0.01$, $d = -1.083$ to $1.033$), and lower confidence versus experts in 9/10 features ($P < 0.05$, $d = -2.076$ to $0.175$). OncoVision showed higher confidence than junior radiologists and comparable levels to senior radiologists, particularly for challenging subclasses such as spiculated masses, isodense lesions, heterogeneously dense tissue, and BI-RADS 4--5 categories. Experts maintained the highest baseline confidence, with only marginal change from AI assistance. Mixed-effects analysis indicated an interaction between reader expertise and AI assistance ($P < 0.001$), with the largest gains observed for junior radiologists and diminishing returns for experts.

AI assistance was associated with higher reader confidence (Fig.~\ref{edfig3}b). Juniors showed the largest gains (from $0.75$--$0.79 \pm 0.07$ to $0.88$--$0.91 \pm 0.02$; $P < 0.001$, $d > 1.3$), followed by seniors ($0.88 \pm 0.06$ to $0.92$--$0.93 \pm 0.02$; $P < 0.001$, $d \approx 0.7$). Experts showed no significant change ($0.93$--$0.96 \pm 0.04$ to $0.94$--$0.96 \pm 0.04$; $P > 0.05$, $d < 0.02$), consistent with their high baseline performance.

\subsubsection{Reading Efficiency}

Diagnostic time was reduced with AI assistance (Fig.~\ref{edfig3}c). Juniors decreased from $38.8$--$41.5 \pm 3.6$--$4.0$ min to $17.0$--$18.6 \pm 1.4$--$1.7$ min per case (55--57\% reduction), seniors from $16.5$--$18.5 \pm 1.4$ min to $6.5$--$8.5 \pm 1.4$--$1.5$ min (54--61\%), and experts from $7.0$--$8.0 \pm 1.1$--$1.2$ min to $5.7$--$6.3 \pm 0.4$ min (18--22\%). AI assistance also reduced variability in reading time (tighter standard deviations), suggesting more predictable workflows across all experience levels. Segmentation time dropped from approximately 120 min to 10--15 min per batch with AI assistance (Fig.~\ref{edfig3}e), showing efficiency gains in annotation-heavy tasks.

\subsubsection{Segmentation Accuracy}

For segmentation, OncoVision achieved higher IoU in mass regions ($0.925$ vs. $0.904$--$0.907$ for radiologists; $P < 0.001$, $d = 0.618$--$0.683$, medium effect) but was slightly less accurate for calcifications ($0.720$ vs. $0.739$--$0.740$; $P < 0.001$, $d = -0.362$ to $-0.377$, small), axilla findings ($0.813$ vs. $0.823$--$0.825$; $P < 0.001$, $d = -0.232$ to $-0.284$, small), and breast tissue ($0.659$ vs. $0.684$; $P < 0.001$, $d = -0.459$ to $-0.476$, small), as shown in Fig.~\ref{edfig3}d. AI predictions showed narrower variance, suggesting higher consistency across cases than human readers.

\section{Limitations and Future Directions}\label{sec:limitations}

Several limitations warrant consideration. The dataset, while geographically diverse, remains limited in size (1,725 mammograms from 500 patients), and class imbalance required augmentation that may introduce synthetic bias. Per-class recall for rare, morphologically complex categories such as spiculated masses and BI-RADS 5 remains low, and sub-visual microcalcifications continue to challenge detection. Radiomic descriptors were derived from predicted masks and were not independently evaluated for stability across acquisition protocols, scanners, or segmentation uncertainty; segmentation errors may therefore propagate into downstream radiomic features. The pipeline does not address architectural distortion, and the reader study was retrospective without washout periods; prospective multi-center trials are needed to confirm diagnostic benefit. Future work will extend the framework to other imaging modalities (ultrasound, MRI, PET), integrate genomic and electronic health record data for personalized risk assessment, incorporate architectural distortion into prediction targets, and leverage federated learning for secure multi-institutional training.

\section{Conclusion}\label{sec:conclusion}

OncoVision advances AI-assisted mammographic interpretation by unifying multi-region segmentation and clinical feature prediction within a single integrated multimodal pipeline, achieving segmentation performance exceeding the nnU-Net baseline across four anatomically distinct ROIs and high discriminative accuracy across ten structured clinical tasks, with the largest gains observed for calcifications a clinically relevant but technically challenging mammographic finding. To the best of our knowledge, existing studies have not reported a unified framework that jointly segments masses, calcifications, axillary findings, and breast tissue while simultaneously predicting a panel of clinical features, including mass morphology, calcification distribution, ACR breast density, and BI-RADS category, thereby addressing a gap left by prior work focused on isolated tasks. The Dependent variant, which employs a dedicated tabular encoder to learn a compact representation of clinical features before fusion with imaging and radiomic representations, outperformed the Independent variant on nine of ten clinical features, making it suitable for deployment in resource-limited settings. A retrospective multi-reader study provided preliminary evidence that OncoVision can improve diagnostic confidence for junior and senior radiologists, reduce reading time by up to 61\%, and achieve segmentation accuracy comparable to or exceeding that of radiologists for mass lesions, supporting its utility as a decision-support tool. However, the dataset, while geographically diverse across urban and rural Bangladesh, remains limited in size, per-class recall for rare, morphologically complex categories remains low, and persistent challenges in detecting sub-visual microcalcifications underscore the need for continued optimization. Future work will extend the framework to other imaging modalities, integrate genomic and electronic health record data for personalized risk assessment, and use federated learning for secure multi-institutional training. By demonstrating that accuracy, interpretability, and practical utility can coexist within a deployable tool, OncoVision represents a step toward more accessible, structured mammographic assessment.

\section*{Acknowledgments}
This study has been funded by North South University (grant CTRG-24-SEPS-20). We are thankful to the expert panel of radiologists that contributed to OncoVision's clinical validation and reader study, which was performed following standard protocols. We would like to acknowledge Dr. Shahnaz Choudhury (Associate Professor, Department of Radiology and Imaging, Bangladesh Medical University), Most. Nahid Farjana (Assistant Professor, Radiology \& Imaging, DNCC Dedicated COVID-19 Hospital), and Sumaiya Akter Shimu (Registrar, Department of Radiology \& Imaging, Popular Medical College Hospital) for their precious clinical input and annotations. Inevaluability of the model’s diagnostic accuracy and clinical relevance was dependent on their knowledge. We also acknowledge the institutions that made data access possible and also to all interdisciplinary team of technicians and clinicians who worked in the technical and clinical implementation of this work.

\section*{Data availability}
The mammography images and associated clinical data used in this study contain protected health information and are subject to institutional data sharing agreements. Due to these ethical restrictions and patient confidentiality requirements, the raw datasets are not publicly available. De-identified data supporting the findings of this study are available from the corresponding author (Dr. Mohammad Monir Uddin) upon reasonable request and with permission from the relevant ethics committees and participating institutions.

\bibliographystyle{unsrtnat}
\bibliography{ref}

\newpage
\onecolumn
\appendix

\section{Inter-Annotator Agreement and Class Balance}
\label{app:annotation}

This appendix provides supporting evidence for the annotation quality and class-balancing procedures described in the Methods. Fig.~\ref{figA1} reports inter-annotator agreement for both segmentation masks and clinical features, along with pre- and post-augmentation class distributions for the ten clinical features and four segmentation regions.

\subsection*{A.1. Pre- and Post-Augmentation Distributional Balance and Inter-Annotator Agreement}

\noindent
\begin{minipage}{\textwidth}
\centering
\includegraphics[width=.80\linewidth,height=0.75\textheight,keepaspectratio]{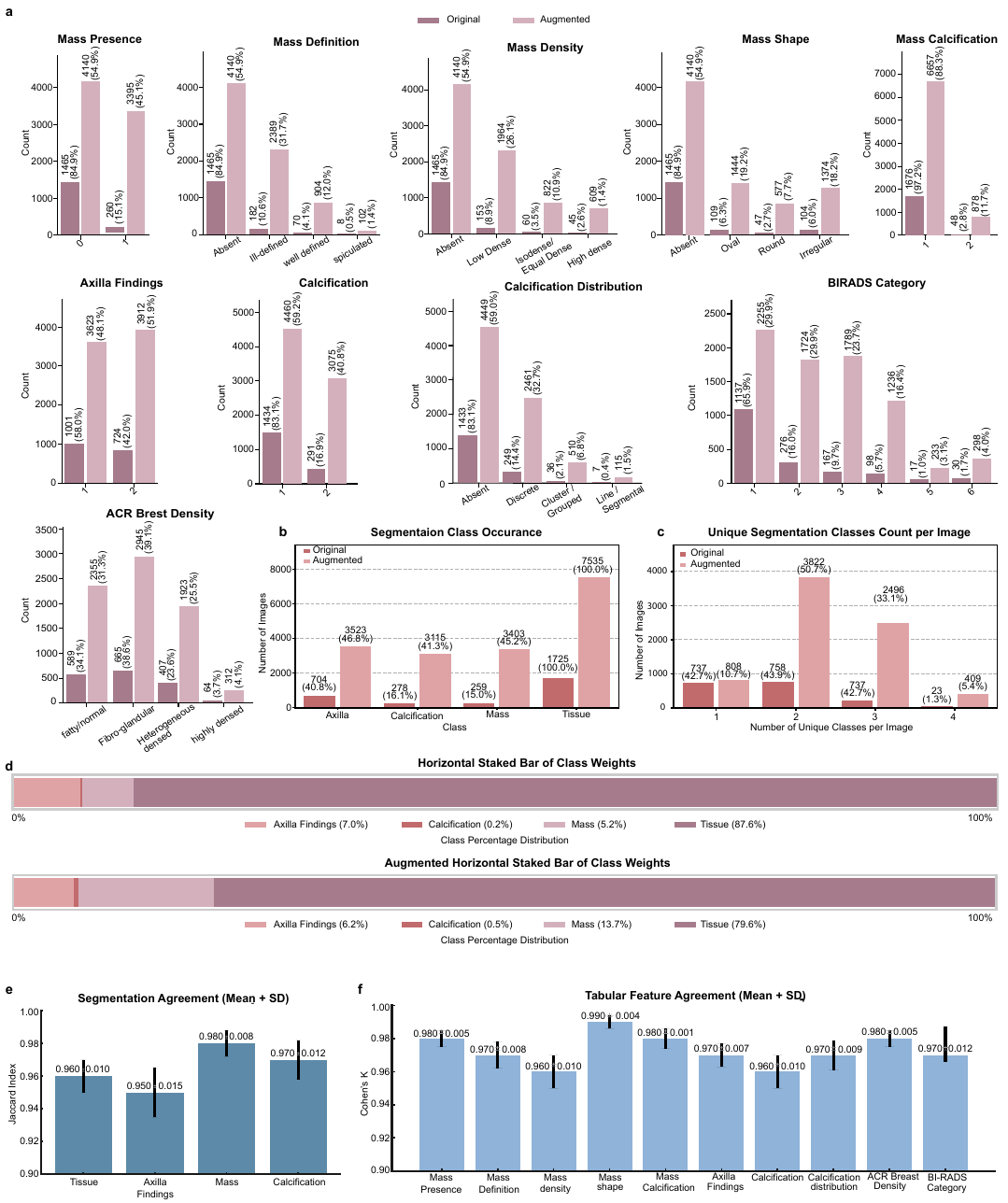}
\captionsetup{hypcap=false}
\captionof{figure}{\textbf{Pre- and post-augmentation distributional balance and inter-annotator agreement.} \textbf{(a)} Class counts and percentages for ten clinical features before and after augmentation, showing improved representation of minority classes such as Mass Definition, Calcification Presence, and BI-RADS Category. \textbf{(b)} Bar plot of segmentation class frequencies across four anatomical regions (Calcification, Axilla Findings, Mass, and Tissue) before and after augmentation. \textbf{(c)} Barplot of unique non-background segmentation classes per image, pre- and post-augmentation, indicating increased multi-class co-occurrence. \textbf{(d)} Stacked horizontal bars of class-weight composition in segmentation masks before and after augmentation, showing redistribution toward balanced representation (e.g., Mass: 5.2\% to 13.7\%; Calcification: 0.2\% to 0.5\%). \textbf{(e)} Mean Jaccard index with standard deviation for inter-annotator segmentation agreement: Tissue (\(0.96 \pm 0.010\)), Axilla Findings (\(0.95 \pm 0.015\)), Mass (\(0.98 \pm 0.008\)), and Calcification (\(0.97 \pm 0.012\)). \textbf{(f)} Mean Cohen's \(\kappa\) with standard deviation for clinical feature annotation agreement across ten diagnostic tasks, ranging from \(0.96 \pm 0.010\) to \(0.99 \pm 0.004\).}
\label{figA1}
\end{minipage}

\clearpage

\section{ROC Analysis of Clinical Feature Prediction}
\label{app:roc}

This appendix provides the per-class ROC curves and AUC values for the ten clinical features, generated from the dependent late-fusion variant. Fig.~\ref{figB1} shows the ROC curves, and Table~\ref{tabB1} reports the corresponding AUC values with 95\% confidence intervals.

\subsection*{B.1. ROC Curves for the Dependent Late-Fusion Variant}

\noindent
\begin{minipage}{\textwidth}
\centering
\includegraphics[width=\linewidth,height=0.75\textheight,keepaspectratio]{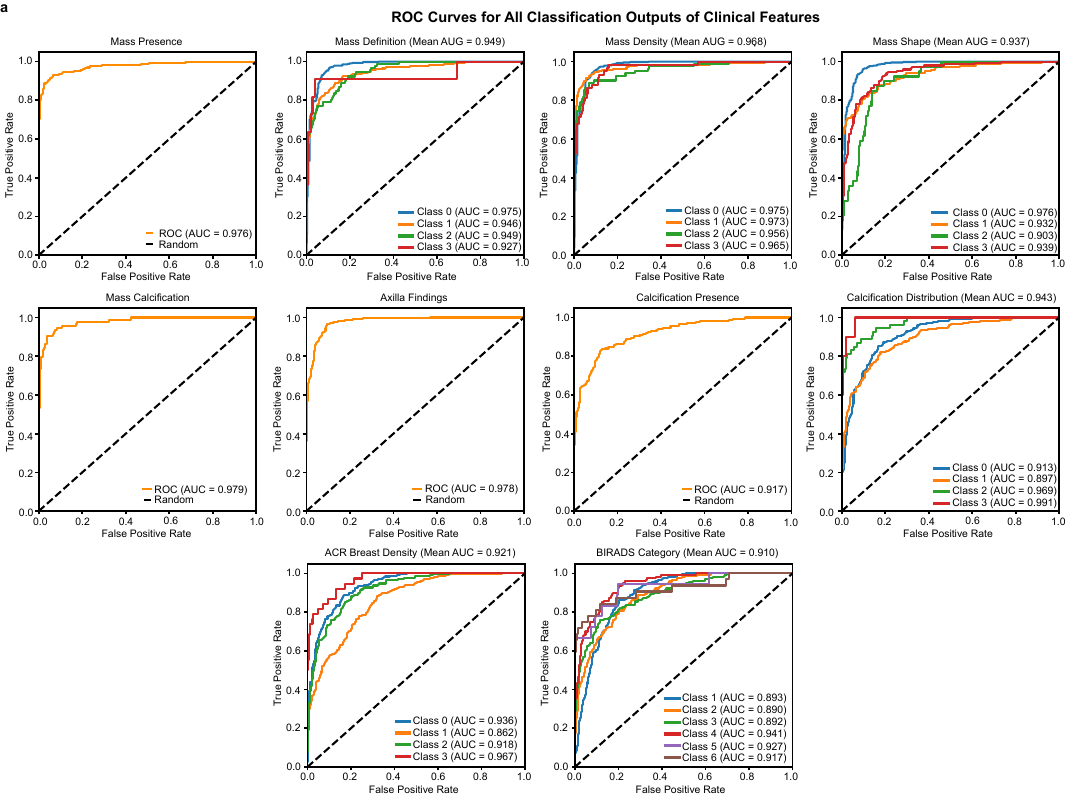}
\captionsetup{hypcap=false}
\captionof{figure}{\textbf{Diagnostic performance across clinical features for the dependent late-fusion variant.} Receiver operating characteristic (ROC) curves for ten clinical diagnostic tasks in breast mammography, including morphological, density, and categorical assessments. Each curve represents the one-versus-rest classification performance for a specific class within a task. Area under the ROC curve (AUC) values are reported with 95\% confidence intervals estimated via 1000 bootstrap samples. The model achieves high discriminative accuracy across all tasks, with AUCs exceeding 0.90 for mass presence, mass definition, mass density, mass shape, mass calcification, axilla findings, and calcification distribution. Performance is robust across subtypes, including spiculated (0.927), irregular (0.939), and segmental (0.991) calcifications, as well as BI-RADS category 4 (0.941) and 5 (0.927). Confidence intervals reflect the large test cohort (\(n = 754\) per task), demonstrating statistical reliability.}
\label{figB1}
\end{minipage}

\clearpage

\subsection*{B.2. AUC Values with 95\% Confidence Intervals}
\vspace{.5cm}
\noindent
\begin{minipage}{\textwidth}
\centering
\small
\captionsetup{hypcap=false}
\captionof{table}{\textbf{AUC values for the ten clinical features with 95\% confidence intervals for the dependent late-fusion variant.} Values are estimated via 1000 bootstrap samples. The mean AUC across all classes is reported where applicable.}
\label{tabB1}
\setlength{\tabcolsep}{6pt}
\renewcommand{\arraystretch}{1.15}
\begin{tabular}{llcccc}
\toprule
\textbf{Feature} & \textbf{Class} & \textbf{Class Name} & \textbf{Mean AUC} & \textbf{95\% CI Lower} & \textbf{95\% CI Upper} \\
\midrule
mass presence & -- & -- & 0.976 & 0.968 & 0.983 \\
\midrule
\multirow{5}{*}{mass definition} & 0 & -- & 0.975 & 0.965 & 0.984 \\
 & 1 & well-defined & 0.946 & 0.928 & 0.962 \\
 & 2 & ill-defined & 0.949 & 0.932 & 0.964 \\
 & 3 & spiculated & 0.927 & 0.905 & 0.947 \\
 & Mean & -- & 0.949 & 0.938 & 0.958 \\
\midrule
\multirow{5}{*}{mass density} & 0 & -- & 0.975 & 0.966 & 0.983 \\
 & 1 & low dense & 0.973 & 0.963 & 0.981 \\
 & 2 & isodense & 0.956 & 0.942 & 0.969 \\
 & 3 & high dense & 0.965 & 0.954 & 0.975 \\
 & Mean & -- & 0.967 & 0.959 & 0.974 \\
\midrule
\multirow{5}{*}{mass shape} & 0 & -- & 0.976 & 0.967 & 0.984 \\
 & 1 & oval & 0.932 & 0.915 & 0.948 \\
 & 2 & round & 0.903 & 0.882 & 0.923 \\
 & 3 & irregular & 0.939 & 0.923 & 0.954 \\
 & Mean & -- & 0.938 & 0.927 & 0.948 \\
\midrule
mass calcification & -- & -- & 0.979 & 0.972 & 0.985 \\
axilla findings & -- & -- & 0.978 & 0.970 & 0.985 \\
calcification presence & -- & -- & 0.917 & 0.898 & 0.934 \\
\midrule
\multirow{5}{*}{calcification distribution} & 0 & -- & 0.913 & 0.892 & 0.932 \\
 & 1 & discrete & 0.897 & 0.874 & 0.918 \\
 & 2 & cluster/grouped & 0.969 & 0.956 & 0.980 \\
 & 3 & line/segmental & 0.991 & 0.985 & 0.995 \\
 & Mean & -- & 0.942 & 0.932 & 0.951 \\
\midrule
\multirow{5}{*}{ACR breast density} & 0 & fatty/normal & 0.936 & 0.920 & 0.951 \\
 & 1 & fibroglandular/mixed & 0.862 & 0.836 & 0.886 \\
 & 2 & heterogeneously dense & 0.918 & 0.900 & 0.935 \\
 & 3 & highly dense & 0.967 & 0.957 & 0.976 \\
 & Mean & -- & 0.921 & 0.911 & 0.930 \\
\midrule
\multirow{7}{*}{BI-RADS category} & 0 & 1 & 0.893 & 0.872 & 0.912 \\
 & 1 & 2 & 0.890 & 0.868 & 0.910 \\
 & 2 & 3 & 0.892 & 0.870 & 0.912 \\
 & 3 & 4 & 0.941 & 0.926 & 0.955 \\
 & 4 & 5 & 0.927 & 0.910 & 0.942 \\
 & 5 & 6 & 0.917 & 0.898 & 0.934 \\
 & Mean & -- & 0.910 & 0.898 & 0.921 \\
\bottomrule
\end{tabular}
\end{minipage}

\clearpage

\section{Radiomic Feature Analysis}
\label{app:radiomics}

This appendix provides the radiomic feature analysis described in the Methods. Fig.~\ref{figC1} reports correlations among the top radiomic features, their association with BI-RADS category, and their distribution across BI-RADS categories for axillary findings.

\subsection*{C.1. Radiomic Feature Correlations and BI-RADS Relationships}

\noindent
\begin{minipage}{\textwidth}
\centering
\includegraphics[width=\linewidth,height=0.75\textheight,keepaspectratio]{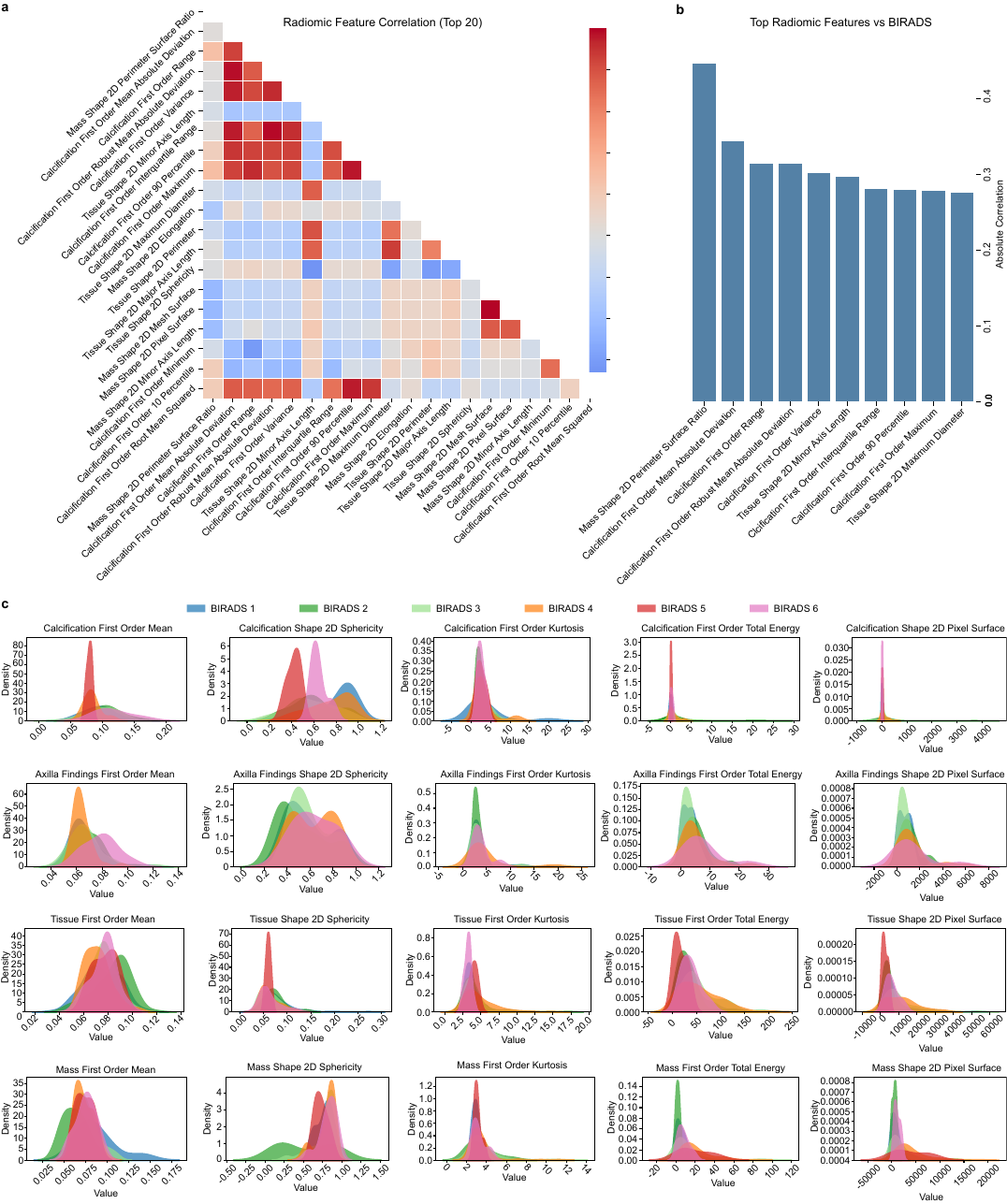}
\captionsetup{hypcap=false}
\captionof{figure}{\textbf{Radiomic feature analysis and BI-RADS relationships.} \textbf{(a)} Heatmap of correlations among the top 20 radiomic features extracted from segmented regions and tabular inputs. Red indicates strong correlations (\(r > 0.80\)), orange indicates moderate correlations, and blue indicates weak correlations. \textbf{(b)} Bar plot showing the absolute correlations of the top 10 radiomic features with the BI-RADS category, ordered from lowest (Tissue Shape 2D Maximum Diameter) to highest (Mass Shape 2D Perimeter Surface Ratio). \textbf{(c)} Kernel Density Estimation (KDE) plots of five radiomic features (first-order mean, sphericity, kurtosis, total energy, and pixel surface) for axilla findings, stratified by BI-RADS categories: BI-RADS~1 (blue), BI-RADS~2 (green), BI-RADS~3 (light green), BI-RADS~4 (orange), BI-RADS~5 (light red), and BI-RADS~6 (pink).}
\label{figC1}
\end{minipage}

\clearpage

\clearpage

\section{Multimodal Overview of Tabular Features and Embedding Alignment}
\label{app:embedding}

This appendix provides a multimodal view of tabular feature distributions across BI-RADS categories and a joint embedding of image-derived and tabular representations before and after augmentation. Augmentation was applied at the patient level to the full dataset before the training/test split. All original and augmented views from a given patient were assigned to the same cohort, ensuring that no augmented samples leaked into the test set. Fig.~\ref{figD1} reports both views.

\subsection*{D.1. Tabular Feature Distribution and Multimodal Embedding by BI-RADS Category}

\noindent
\begin{minipage}{\textwidth}
\centering
\includegraphics[width=.70\linewidth,height=1\textheight,keepaspectratio]{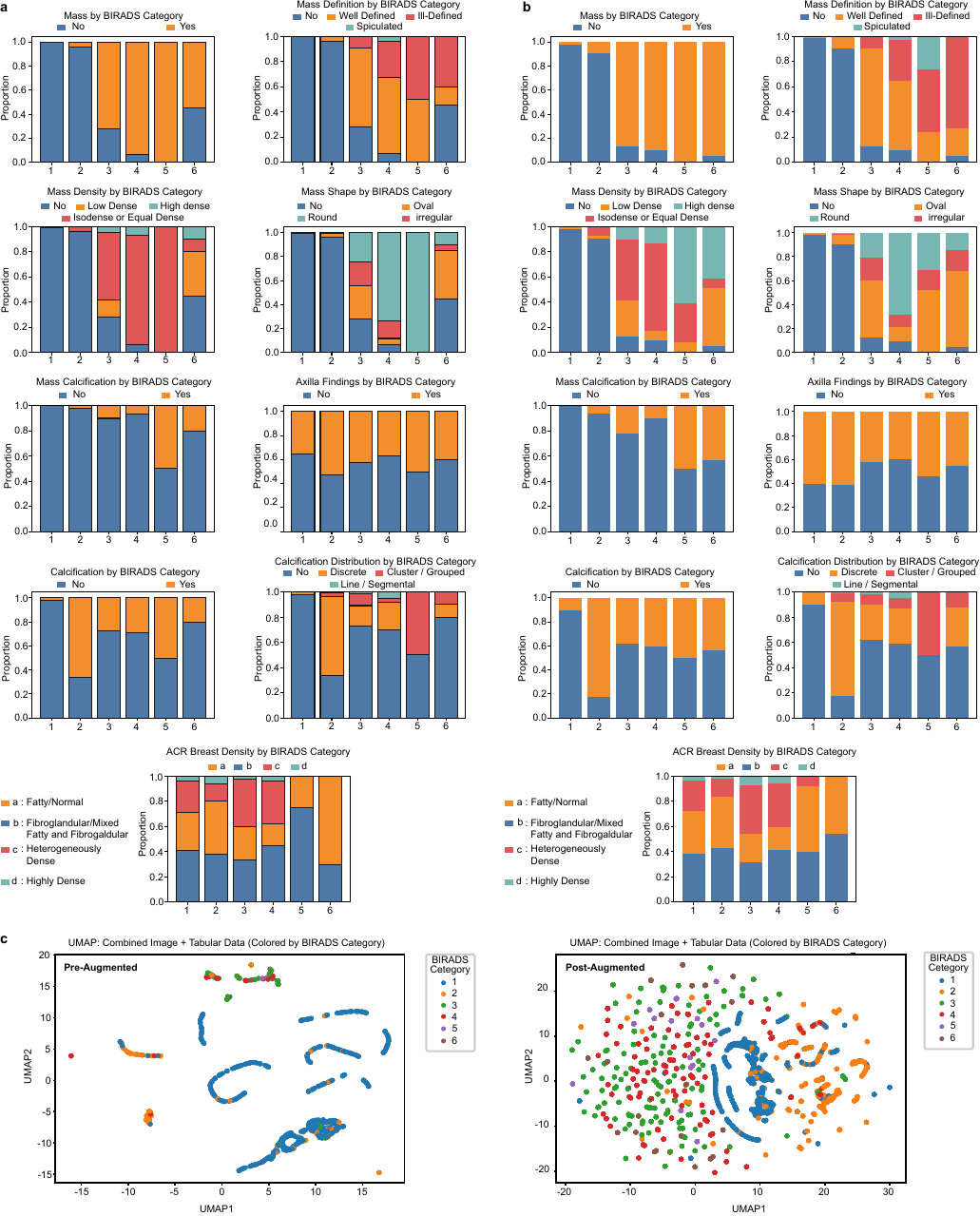}
\captionsetup{hypcap=false}
\captionof{figure}{\textbf{Multimodal overview of tabular features by BI-RADS category and embedding alignment before and after augmentation.} \textbf{(a-b)} Grouped stacked bar plots showing the distribution of nine tabular features across BI-RADS categories, before (a) and after (b) augmentation. Stacked bars represent the proportion of subcategories within each BI-RADS class. Pre-augmentation plots reveal inherent imbalances and skewed representations; post-augmentation plots show more uniform feature representation, mitigating potential bias in downstream training. \textbf{(c)} Multimodal UMAP embedding of combined image-derived and tabular features, colored by BI-RADS category, comparing pre- and post-augmentation datasets. Pre-augmentation embeddings show natural clustering by BI-RADS category; post-augmentation embeddings show enhanced density and overlap, indicating improved balance and representation of underrepresented categories. Together, these panels demonstrate that augmentation preserves intrinsic relationships while reducing class imbalance.}
\label{figD1}
\end{minipage}

\newpage
\subsection*{E.1. Web Platform}
\vspace{.5cm}

This appendix provides a screenshot of the OncoVision web platform, which has been deployed at a partner hospital for real-time clinical use. The platform supports interpretation of mammographic studies, from image upload to structured report generation, and integrates directly into radiologist workflows. It produces interpretable outputs, including segmentation overlays, structured clinical predictions with confidence scores, and Grad-CAM visualizations, allowing clinicians to review the model's reasoning alongside the original images. The interface also serves as a teaching tool for medical students and residents by exposing the model's decision process in an accessible visual format.

The radiologist uploads a multiview patient mammogram, comprising cranio-caudal and mediolateral oblique views of both breasts, in DICOM or standard image format. The platform then runs the OncoVision inference pipeline: segmentation masks are overlaid on each view, ten structured clinical features are predicted with confidence scores, and Grad-CAM heatmaps are generated across early, mid, and bottleneck layers. Per-view predictions are aggregated into a single structured report, which is compiled into a downloadable PDF for clinical review.

A key feature is the aggregation of predictions across views. The model is applied independently to each view, producing per-view segmentation masks and clinical feature predictions with confidence scores, which are then combined into a patient-level report. For ordinal features such as BI-RADS category, the highest-risk prediction across views is retained, so that a single suspicious view is not masked by otherwise benign findings. For example, if one view shows no mass and yields a BI-RADS category of 1 or 2, while another view shows a mass and yields a BI-RADS category of 3 to 6, the higher category is selected. Other features are aggregated using either the maximum value or the highest-confidence prediction, depending on the feature type. This approach ensures that the final report reflects the most clinically relevant finding across the entire study while preserving per-view predictions for review.

\vspace{.5cm}
\noindent
\begin{minipage}{\textwidth}
\centering
\includegraphics[width=\linewidth,height=0.75\textheight,keepaspectratio]{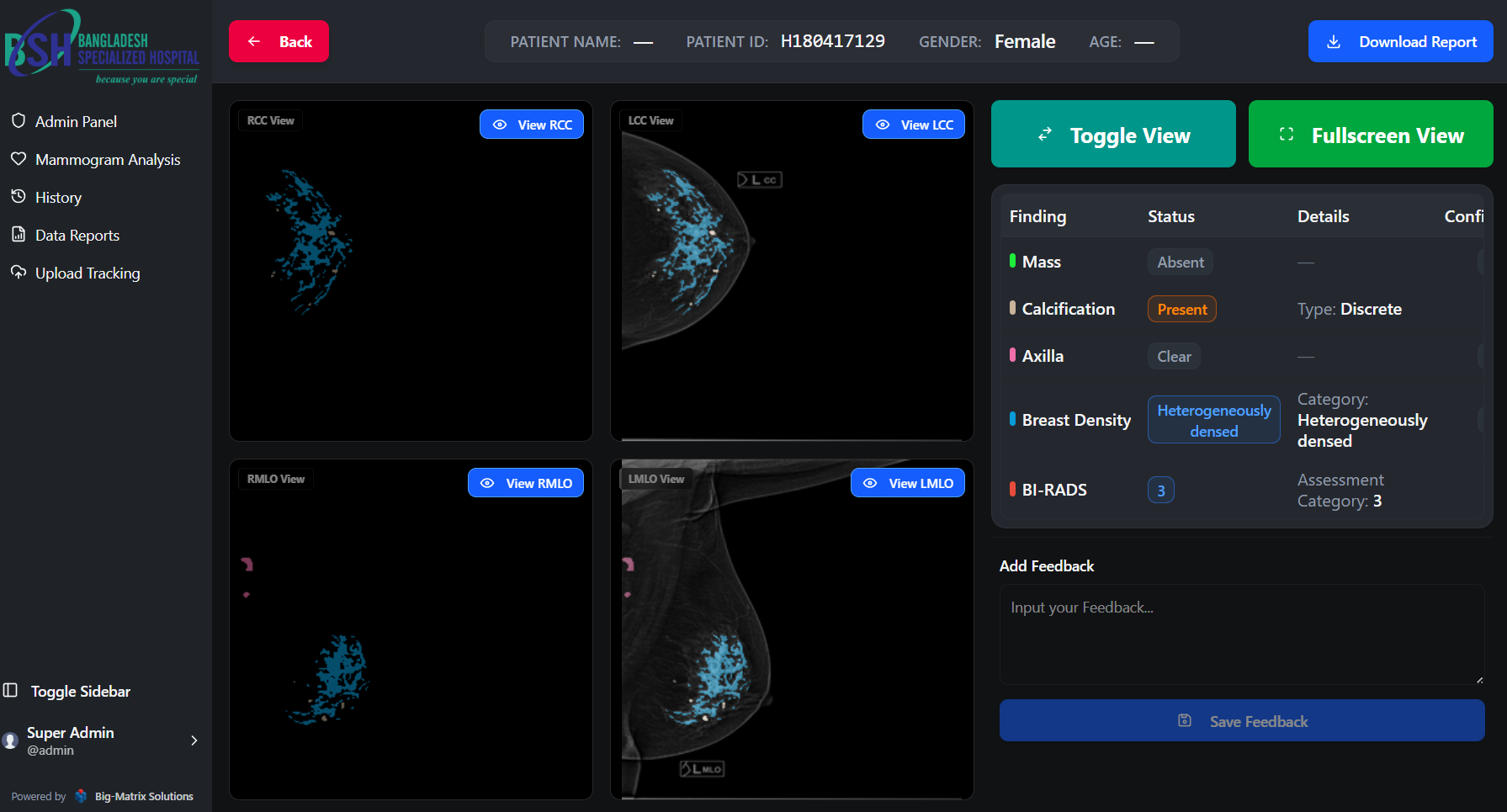}
\captionsetup{hypcap=false}
\captionof{figure}{\textbf{Screenshot of the web platform processing a bilateral mammographic study.} Each image displays high-resolution segmentation masks for mass (green), calcification (cream), axilla findings (pink), and breast tissue (blue), overlaid on the original mammogram with interactive zoom, hover, and pan via WebGL. The right sidebar presents predictions from the dependent late-fusion variant for ten clinical features with confidence scores, including mass presence, definition, density, shape, calcification status and distribution, axilla findings, ACR breast density, and BI-RADS category. A downloadable PDF report compiles all outputs, including per-image visualizations, structured predictions, Grad-CAM heatmaps (toggleable across early, mid, and bottleneck layers) that highlight regions driving predictions, such as lesion boundaries and microcalcification clusters, and metadata, ensuring clinical traceability and integration.}
\label{figE1}
\end{minipage}

\end{document}